\documentclass[10pt]{article}
\usepackage{geometry}
\usepackage{graphicx}
\usepackage[FIGTOPCAP,nooneline]{subfigure}
\usepackage[T1]{fontenc}
\usepackage{lmodern}
\usepackage{bm}
\usepackage{amsmath, amsfonts, amssymb}
\usepackage{cite}
\usepackage{multirow}
\usepackage{afterpage, array, rotating}
\usepackage{url}
\usepackage{authblk}
\usepackage{color}
\usepackage{algorithm}
\usepackage[noend]{algorithmic}
\usepackage{bm}
\usepackage{mathtools}
\usepackage{booktabs}
\usepackage{tabularx}
\usepackage{multirow}
\usepackage{rotating}
\usepackage{lineno}

\newcolumntype{A}{>{\centering\arraybackslash}p{5cm}}
\newcolumntype{B}{>{\centering\arraybackslash}p{1.2cm}}
\newcolumntype{C}{>{\centering\arraybackslash}p{3cm}}
\newcolumntype{D}{>{\raggedright\arraybackslash}p{11cm}}
\newcolumntype{E}{>{\raggedleft\arraybackslash}p{1.2cm}}

\title{
Large Language Model Turnover Undermines Screening for Artificial Intelligence-Assisted Scientific Writing}

\author[1]{Kazuki Nakajima}
\author[2]{Takayuki Mizuno}

\affil[1]{Tokyo Metropolitan University}
\affil[2]{National Institute of Informatics}

\begin{document}
\date{}
\maketitle

\begin{abstract}
Journals and conferences have begun to screen submitted manuscripts for text written using large language models (LLMs).
The reliability of this screening rests on benchmark evaluations against a fixed set of LLM versions, while the versions in actual use keep changing.
Here we quantify how this LLM turnover affects the screening of scientific manuscripts.
We paired 4,000 pre-ChatGPT abstracts from the Proceedings of the National Academy of Sciences with their rewrites by 23 LLM versions from three vendors, released between June 2023 and August 2026.
We then trained detectors under maintenance scenarios ranging from a detector retrained on every new version to one trained once and never updated.
Detectors trained only on a vendor's past versions can collapse at the boundaries between model generations: calibrated to falsely flag 1\% of human-written abstracts, they catch above 99\% of rewrites just before the sharpest boundary and 3.8\% just after it.
Detectors trained on later versions can also miss rewrites of earlier ones.
Vocabulary differences between versions largely track where detection transfers and where it fails.
In the two screening scenarios we simulated, screens covering all 23 versions either flagged one in eight human-written abstracts or missed one in three rewrites of the newest version.
Indeed, a commercial detector missed most rewrites of the version just after the sharpest boundary while flagging almost no human-written abstracts.
Research-integrity policy should therefore treat the benchmark accuracy of a detector as provisional, to be re-verified with every LLM release, including earlier versions.
\end{abstract}

\section{Introduction}

Large language models (LLMs) have entered the process of scientific research, from the search and synthesis of prior literature~\cite{asai2026literature} to the drafting and editing of manuscripts~\cite{liu2024detectability,yu2025cheat}, the writing of peer review reports~\cite{liang2024monitoring}, and the preparation of grant proposals~\cite{qian2026funding}.
LLM assistance is already widespread, with estimates of LLM-modified content reaching up to 22\% in computer science papers by September 2024~\cite{liang2025quantifying} and at least 13.5\% of biomedical abstracts published in 2024, rising to about 20\% for some countries and fields and about 40\% for some of their combinations~\cite{kobak2025delving}.
Analyses by publishers likewise report a rapid rise in submitted manuscripts and peer review reports containing text produced by an LLM~\cite{naddaf2026howmuch}.
Journals, conferences, and editorial organizations ask authors to disclose such use~\cite{icmje2026,pnas2026authors,lu2026neurips}, and the estimates above suggest that the use has spread faster than its disclosure~\cite{naddaf2025detects}.
These LLMs accelerate scientific production and lower language barriers, while they weaken the signals, such as the complexity of the writing, that readers have used to judge scientific work~\cite{kusumegi2025production}.
How science should respond to this trade is debated across disciplines~\cite{binz2025practice,kusumegi2025production}.
Journals and conferences face it one submission at a time, ahead of any resolution of that debate.

Journals and conferences have begun to screen individual submissions with automated detectors that estimate how much of a text was written or edited with an LLM.
A leading conference in computer science analyzed every submission to one of its tracks with a commercial detector and rejected 178 of them without review in 2026, which amounted to 18.4\% of the submissions to that track~\cite{lu2026neurips}.
Similarly, a journal published by the National Academy of Sciences lists the same product among the tools it uses to assess submissions~\cite{pnasnexus2026screening}.
Guidance for medical journal editors now devotes a dedicated section to the use of AI in publishing~\cite{icmje2026}.
Such detectors usually express this estimate as a score and return a verdict once the score passes a threshold~\cite{emi2024pangram,lu2026neurips}.
The two kinds of mistakes such a verdict can make fall on different researchers.
When a manuscript written by its authors is judged to involve an LLM, the authors may carry the burden of contesting the verdict and a rejection of the manuscript.
When a manuscript largely written with an LLM passes as human writing, and the use was neither disclosed nor permitted, the burden can fall on reviewers and editors who assume the text reflects its authors.
In practice, deployed detectors are often evaluated at a fixed low rate of flagging text that a human wrote~\cite{krishna2023paraphrasing,tufts2025practical,yu2026peerreview}, yet the flags that remain have been reported to fall unevenly~\cite{liang2023biased,park2026style}.

The reliability of screening scientific manuscripts rests on benchmarks assembled from the LLM versions available when a detector is evaluated~\cite{li2024mage,wang2024m4,dugan2024raid,wu2024detectrl,yu2025cheat,liu2024detectability,liu2025generalization}.
Detectors can lose their accuracy on LLMs absent from their training data~\cite{pu2023deepfake,tufts2025practical,antoun2024text,wu2025survey}.
This loss extends to successive LLM versions within a single LLM family~\cite{yu2025evobench}.
In the scientific literature, the period over which a released LLM is actively used has shortened with each release year~\cite{trisovic2026lifespan}.
A detector in service can be asked to judge a version that it was not built against, and how to adapt a detector to newly arriving models is an active line of work~\cite{guo2024detective,li2025continual,sun2026ridgeft,ren2026moving}.
How this LLM turnover affects the screening of scientific manuscripts remains to be understood under the conditions a screen faces in service, where retraining takes effort and flags of human-written text must stay rare.

Here we quantify how detection of scientific abstracts rewritten with an LLM behaves as newer LLM versions arrive.
We paired 4,000 abstracts published before the release of ChatGPT~\cite{openai2022chatgpt} with their rewrites by 23 LLM versions released by three vendors between June 2023 and August 2026.
We then trained detectors under maintenance scenarios that mirror practice, from retraining on every new version to training a detector once and never updating it.
A detector trained on the version it screens caught almost all of its rewrites, whereas the same detector could miss most rewrites of a later generation from the same vendor.
These drops concentrated at a few transitions between versions.
The vocabulary difference between the rewrites of two versions largely tracks where detection transferred and where it failed.
The two screening scenarios we simulated divided their errors differently, one flagging more human-written abstracts and the other missing more rewrites.
A commercial detector deployed for screening shows the same pattern, missing most rewrites of the version at the sharpest of these transitions while flagging almost no human-written abstracts.
Together, our results indicate that the reliability of screening is not a fixed property of a detector, and that it has to be re-earned as LLM versions turn over.

\section{Results}

\subsection{Detection across LLM versions}

For scientific abstracts, we use 34,925 articles spanning four disciplinary domains (i.e., Health Sciences, Life Sciences, Physical Sciences, and Social Sciences) published between 2009 and 2018 in the Proceedings of the National Academy of Sciences of the United States of America (PNAS) (see Methods).
For LLMs, we focus on 23 versions from three vendors: OpenAI (14 versions), Meta (4 versions), and Alibaba (5 versions), released between June 2023 and August 2026 (see Methods and Supplementary Table~S1).
We draw 4,000 articles equally from the four disciplinary domains, and each LLM version rewrote their abstracts, which yielded 91,953 retained rewrites (see Methods).
Throughout, we use `AI-generated text' for any text that an LLM produced, and `AI-rewritten' for the specific form studied here, in which an LLM rewrites an existing human-written abstract.

We examine whether detectors can tell AI-rewritten abstracts apart from the original human-written abstracts.
We train and evaluate detectors on the 3,955 of the 4,000 papers whose rewrites were retained for all 23 versions (see Methods).
We calibrate each detector's threshold on abstracts from the 30,925 papers that were never rewritten and that no detector sees in training, to flag 1\% of them as AI-rewritten; this `false positive rate' of 1\% is our primary operating point (see Methods).

We first ask whether publicly released non-commercial detectors can catch these rewrites.
Four such detectors, used without any training on our corpus (i.e., the GPT-2 output detector of OpenAI~\cite{solaiman2019}, RADAR~\cite{hu2023radar}, Fast-DetectGPT~\cite{bao2024fastdetectgpt}, and Binoculars~\cite{hans2024binoculars}), missed more than two-thirds of the rewrites of every version (Supplementary Section~S5).
These detectors also ordered AI-rewritten and human-written abstracts only modestly better than chance (Supplementary Section~S5), which indicates that the misses do not come from our operating point alone.
Similarly low rates for such detectors have been reported on scientific text~\cite{liang2025quantifying} and at fixed low false positive rates on texts and models outside their training data~\cite{tufts2025practical}.
We therefore train detectors on our corpus.
These detectors learn from the same domain of text that they will later judge, a setting that favors detection.
Any collapse we still find despite this advantage could be serious in practice.

We simulate detection under four maintenance scenarios, which specify which versions' rewrites are used for training when a detector judges the rewrites of a target version $v$. 
In the ``current version'' scenario, the detector trains on rewrites of $v$ itself; this idealizes a detector built with full access to the version being screened. 
In the ``past and current versions'' scenario, it trains on rewrites of $v$ and of all earlier versions from the same vendor; this represents a detector retrained the moment a new version appears. 
In the ``past versions only'' scenario, it trains on the earlier versions from the same vendor alone; this reflects the common situation that a new version's text arrives before any training data for that version exist. 
In the ``frozen at a specific version'' scenario, it trains on one designated version only and is never updated; in our experiments, the designated version is the earliest version we study from each vendor. 

\begin{figure}[p]
\centering
\includegraphics[width=1\linewidth]{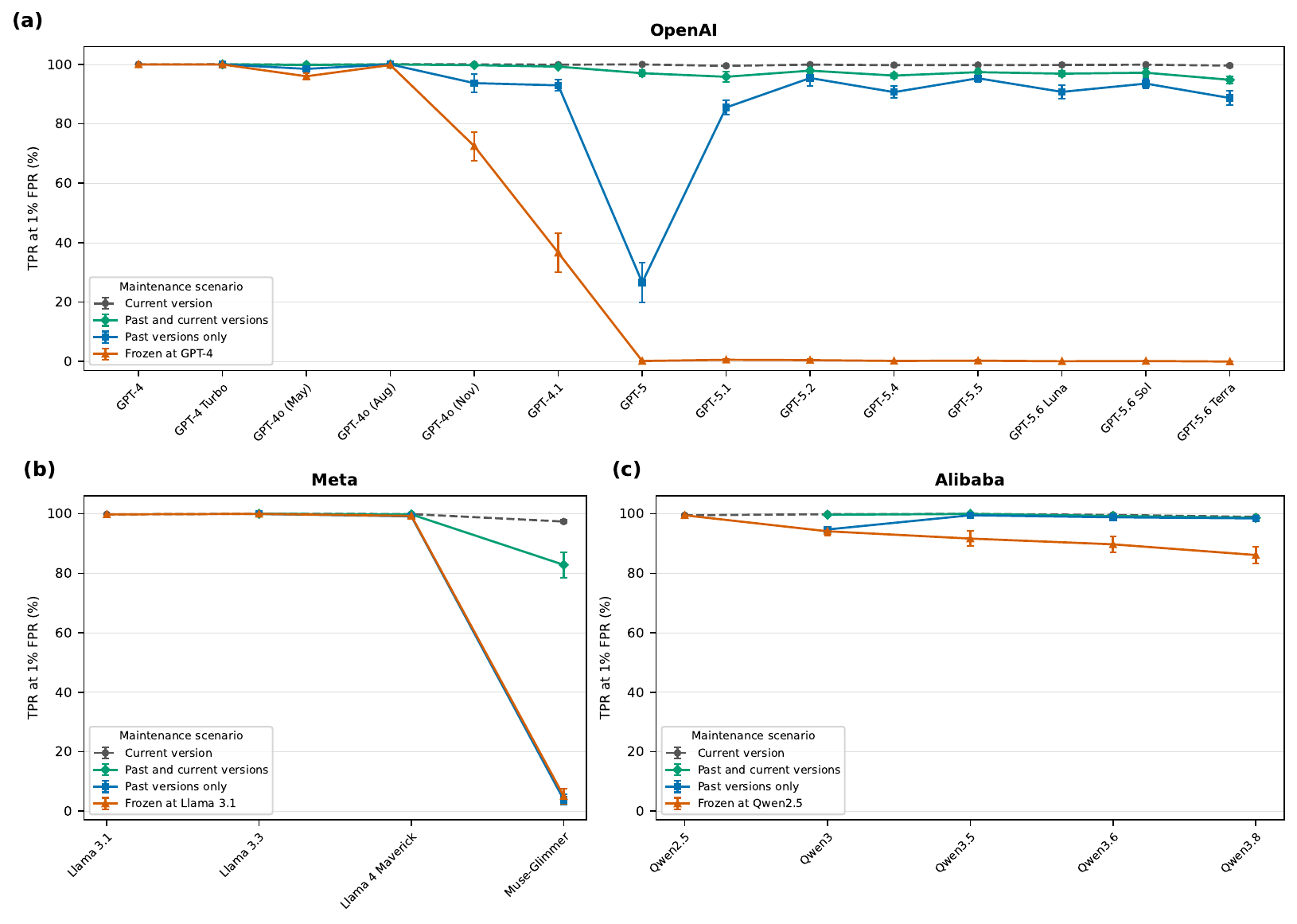}
\caption{
Detection across the 23 LLM versions under four maintenance scenarios. 
Each panel shows one vendor's versions in ascending order of release date. 
The vertical axis is the share of each version's rewrites that the detector catches at the 1\% false positive operating point (TPR at 1\% FPR; see Methods). 
Detectors are trained on rewrites of the version itself (``Current version''), of the version and all earlier versions from the same vendor (``Past and current versions''), of the earlier versions alone (``Past versions only''), or of the earliest version we study from the vendor, never updated (``Frozen at GPT-4''; frozen at Llama 3.1 and Qwen2.5 in the other panels). 
Points are means and error bars are standard deviations over 10 random training-evaluation splits. 
}
\label{fig:1}
\end{figure}

For each of the 23 LLM versions, we train detectors under the four maintenance scenarios and measure the share of its rewrites that each detector catches at the 1\% operating point~\cite{krishna2023paraphrasing,tufts2025practical} (i.e., the true positive rate, or TPR; see Methods).
When the detector has trained on rewrites of the very version it judges, detection is nearly perfect (i.e., the median TPR across the 23 versions is 99.9\%; see ``Current version'' for each vendor in Fig.~\ref{fig:1}). 
Retraining on the current and all earlier versions performs almost as well (the median TPR across the 23 versions is 99.3\%; ``Past and current versions'' for each vendor in Fig.~\ref{fig:1}).
However, training only on earlier versions, the common situation in practice, behaves differently (see ``Past versions only'' for each vendor in Fig.~\ref{fig:1}). 
For 18 of the 20 versions that have earlier versions to train on, detection stays above 85\% (the median TPR across these 18 versions is 95.4\%; the lowest is 85.5\%, for GPT-5.1).
It collapses twice, to $26.6\%$ for GPT-5, the first version of the GPT-5 generation, and to $3.8\%$ for Muse-Glimmer, the first open-weight model of Meta's Muse family~\cite{meta2026glimmer}.
Other transitions survive, including Qwen's move to the Qwen3 architecture (94.7\%) and Meta's move to the mixture-of-experts architecture of Llama 4 (99.2\%).
A detector frozen at the earliest version we study from each vendor catches far fewer of the rewrites of later versions, along a trajectory that differs across vendors and does not decline monotonically.
Frozen at GPT-4, it catches at most 0.6\% of the rewrites of any version from GPT-5 onward; frozen at Qwen2.5, it still catches 86\% of the rewrites of Qwen3.8 (the ``Frozen at'' lines in Fig.~\ref{fig:1}).
These results are partially consistent with previous studies~\cite{pu2023deepfake,li2024mage,dugan2024raid,yu2025evobench} in that detectors of AI-generated text can lose accuracy on text generated by LLMs absent from their training data.
The collapse at generational boundaries persisted with a word-frequency detector that involves no neural network~\cite{liang2024monitoring}, at operating points of 5\% and 0.5\% FPR, and with two alternative rewriting prompts~\cite{chemaya2024perceptions} (Supplementary Section~S7).

The four maintenance scenarios keep the training data and the judged version within one vendor. 
In practice, a detector trained on one vendor's rewrites will also face text produced by other vendors' models. 
We therefore train one detector per version and test each on every version, which yields all $23 \times 23$ train-to-test detector pairs.

\begin{figure}[p]
\centering
\includegraphics[width=0.7\linewidth]{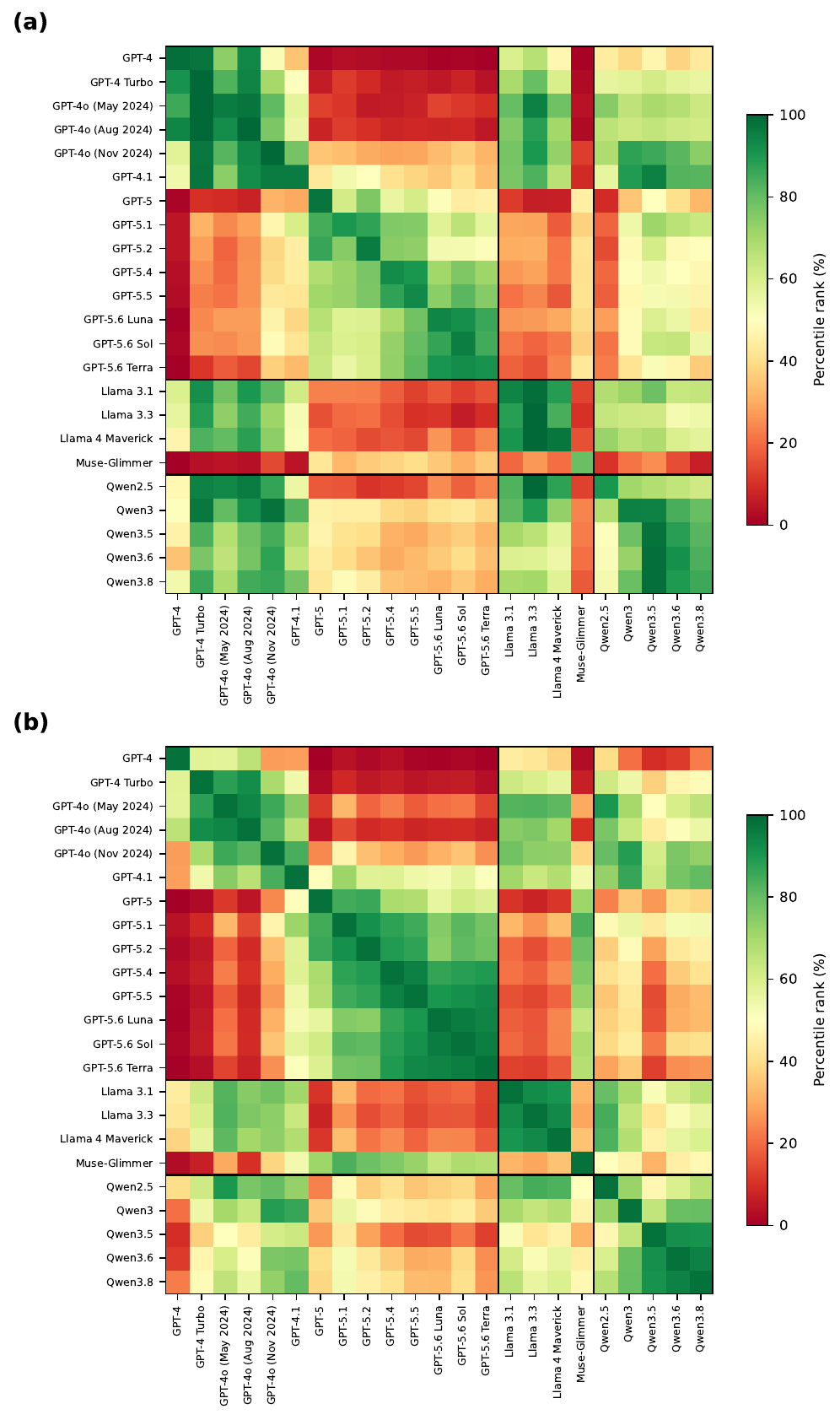}
\caption{
Cross-version detection and vocabulary closeness for all pairs of the 23 LLM versions.
(a) Each cell shows the TPR at 1\% FPR of the detector trained on the rewrites of the row version and evaluated on the rewrites of the column version, averaged over the 10 random splits.
(b) Each cell shows the closeness between the rewrites of the row version and the rewrites of the column version, measured by the JSD.
In both panels, each cell is colored by the percentile rank of its value among all $23 \times 23$ cells of that panel.
In panel (a), a rank of 100 means the highest TPR.
In panel (b), a rank of 100 means the smallest JSD.
}
\label{fig:2}
\end{figure}

Figure~\ref{fig:2}(a) shows the TPR at 1\% FPR of the detector trained on rewrites of version $i$ and tested on the rewrites of version $j$ in the entry for a pair $(i, j)$. 
Because these values crowd near 100\% for many pairs, the color scale shows their percentile ranks, which makes the relative structure visible. 
Detection carries over unevenly, and the pattern follows model generations rather than vendors.
A detector trained on one vendor's version often catches a contemporary version of another vendor; the detector trained on GPT-4o (November 2024) catches 99.5\% of the rewrites of Llama 3.3.
A detector can fail across generations of its own vendor; the detector trained on GPT-4 catches 0.2\% of the rewrites of GPT-5.4.
The failure also runs in the opposite direction; the detectors trained on the three latest OpenAI versions, GPT-5.6 Luna, Sol, and Terra, catch at most 0.2\% of the rewrites of GPT-4.
These results are partially consistent with a previous study~\cite{antoun2024text} in that detection can transfer between different LLMs.

Why does a detector trained on one version catch the rewrites of some other versions and miss those of others? 
One candidate explanation is vocabulary; word-level distributions are known to differ across LLMs~\cite{sun2025idiosyncrasies}, and if two versions favor the same words, a detector trained on one may catch the other.
We therefore measure how far apart the vocabularies of two versions' rewrites are. 
For each version, we pool its retained rewrites, lowercase the text, extract tokens as maximal runs of alphanumeric characters, and discard tokens that consist only of digits. 
This yields one word-frequency distribution per version, normalized to relative frequencies. 
For every pair of versions, we compute the Jensen--Shannon divergence (JSD)~\cite{lin1991divergence} between the two distributions $P$ and $Q$, a standard measure of corpus similarity in natural language processing~\cite{plank2011effective}, defined as $\mathrm{JSD}(P, Q) = \frac{1}{2} D_{\mathrm{KL}}(P \| M) + \frac{1}{2} D_{\mathrm{KL}}(Q \| M)$ with $M = \frac{1}{2}(P + Q)$, where $D_{\mathrm{KL}}$ denotes the Kullback--Leibler divergence with logarithms taken to base 2. 
The JSD ranges from 0 for identical distributions to 1 for distributions with no words in common.

The JSD forms a similar pattern (see Fig.~\ref{fig:2}(b)). 
For each of the 253 unordered pairs of versions, we compare the pair's JSD with the average of its two TPRs at 1\% FPR, one for each train-to-test direction. 
The larger the JSD, the lower this averaged TPR (Spearman rank correlation $\rho = -0.85$, $p < 0.001$; see Supplementary Section~S6). 
The association holds both within and across vendors, and remains after adjustment for the elapsed time between releases (see Supplementary Section~S6).
It also held with the word-frequency detector, at the other operating points, and with the alternative prompts (Supplementary Section~S7).
The two versions farthest from their nearest predecessor from the same vendor by JSD (Muse-Glimmer and GPT-5) are exactly the two versions with the lowest past-only TPR. 

Taken together, detectors trained only on earlier versions worked for most new versions, and their largest failures coincided with the largest vocabulary shifts between successive versions.
This raises the question of whether an institution can keep up with the LLM turnover by collecting rewrites of every new LLM version and retraining its own detectors.

\subsection{Trade-off between catching AI-rewritten abstracts and flagging human-written abstracts}

The results for the maintenance scenarios highlight that detection can collapse when an institution's training data lag behind the newest LLM versions.
We now focus on what happens when the institution holds the rewrites of all 23 LLM versions.
We consider two screening scenarios that use them at once.
In the union scenario, the institution runs the 23 version-specific detectors side by side, each at its own threshold of 1\% FPR, and flags an abstract when at least one detector fires.
In the pooled scenario, the institution trains a single detector on the rewrites of all 23 LLM versions pooled together and sets its threshold in the same way as for the other detectors.
For each scenario, we examine both the share of human-written abstracts it flags and the share of AI-rewritten abstracts it misses.

The union scenario catches nearly every AI-rewritten abstract.
Even for Muse-Glimmer, the union catches 98.6\% of the AI-rewritten abstracts on average, and more for every other version.
The cost falls on the human side.
Each of the 23 detectors alone holds its false positive rate at 1\%.
The abstracts falsely flagged by different detectors only partly overlap, and the false flags therefore accumulate across the 23 detectors.
Over the 10 random splits, the union flags 11.9\% $\pm$ 1.2\% of the held-out human-written abstracts, that is, of the 791 original abstracts per split that no detector saw in training.
An institution that screens with the union would thus falsely flag roughly one in eight genuinely human abstracts.

The pooled scenario rarely flags human-written abstracts.
The single detector flags 0.8\% $\pm$ 0.3\% of the held-out human-written abstracts over the 10 random splits, close to the 1\% at which its threshold is set.
For most LLM versions, the pooled detector also catches almost every AI-rewritten abstract, with a median miss rate of 2.8\% across the 23 versions.
The misses concentrate on a few LLM versions.
The miss rate reaches 33.8\% $\pm$ 3.4\% for Muse-Glimmer, followed by 8.8\% for Qwen2.5 and 6.6\% for GPT-5.6 Terra.
The pooled detector trains on the same number of rewrites as every other detector in this study, one per training paper drawn from the 23 versions.
The pooled detector misses one in three Muse-Glimmer rewrites even though its training data contain Muse-Glimmer rewrites.

In practice, LLM versions arrive one after another, and a screen at any moment can contain only the versions released up to that moment.
To simulate this growth, we sort the 23 LLM versions by release date and build, for each $k$ from 1 to 23, a union of the first $k$ version-specific detectors and a pooled detector trained on the rewrites of the first $k$ versions.
The screens with $k = 23$ are the union and pooled scenarios above, and a screen with a smaller $k$ is the same screen at an earlier moment.

\begin{figure}[t]
\centering
\includegraphics[width=1\linewidth]{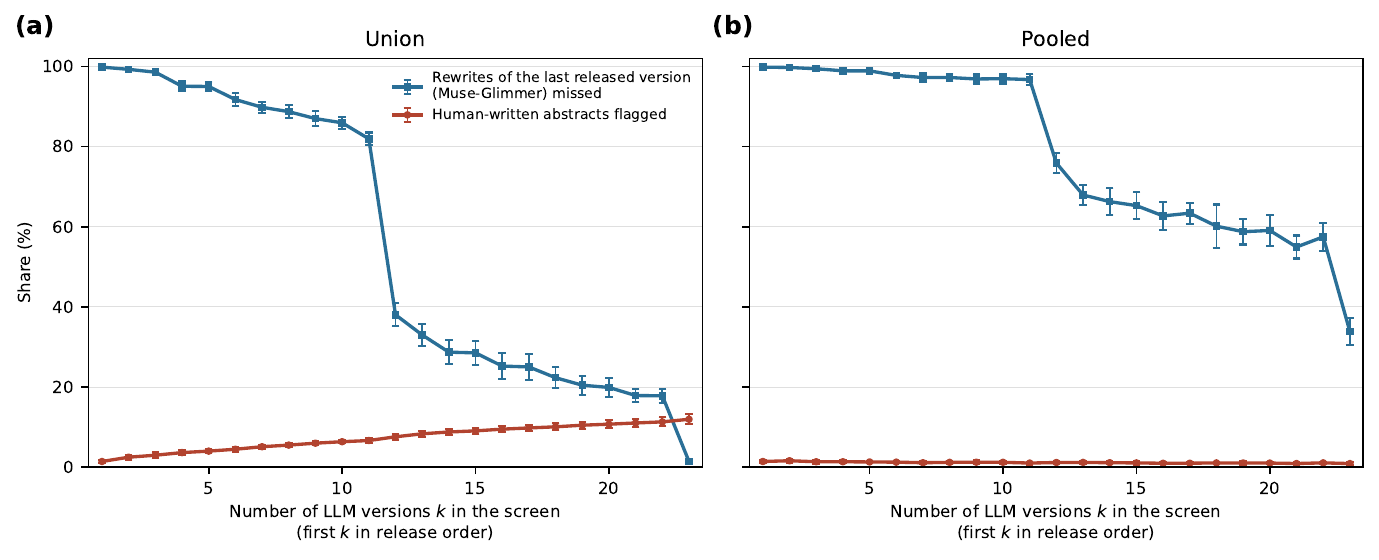}
\caption{
Two screening scenarios built from the first $k$ LLM versions in release order.
(a) Union scenario. (b) Pooled scenario.
For each $k$, the blue curve shows the share of the rewrites of the last released LLM version, Muse-Glimmer, that the screen misses, and the red curve shows the share of the held-out human-written abstracts that the screen flags.
Each point is the mean over the 10 random splits, and error bars show standard deviations.
}
\label{fig:3}
\end{figure}

Figure~\ref{fig:3} shows, for each $k$, the share of held-out human-written abstracts flagged and the share of Muse-Glimmer rewrites missed, where Muse-Glimmer is the last released of the 23 LLM versions.
In the union scenario (Fig.~\ref{fig:3}(a)), every added detector adds false flags, and the flagged share climbs from 1.4\% at $k = 1$ to 11.9\% at $k = 23$.
The missed share of the Muse-Glimmer rewrites falls as $k$ grows and reaches 17.8\% at $k = 22$.
At $k = 23$, when the detector trained on the Muse-Glimmer rewrites joins the union, the missed share falls to 1.4\%.
In the pooled scenario (Fig.~\ref{fig:3}(b)), the flagged share stays near 1\% at every $k$.
The missed share of the Muse-Glimmer rewrites stays above 50\% for pools of 12 to 22 versions.
The same holds at the GPT-5 boundary: trained on the 11 versions released before GPT-5, the pooled detector catches 26.7\% of the GPT-5 rewrites, as few as the detector trained on OpenAI's earlier versions alone (26.6\%; Fig.~\ref{fig:1}).
At $k = 23$, when the Muse-Glimmer rewrites enter the pool, the missed share falls to 33.8\%.
Training the pooled detector on all of the available rewrites of the pooled versions, rather than on one rewrite per paper, with the optimizer, learning rate, and number of epochs of the main text, did not remove this miss rate: it still missed 44\% to 67\% of the Muse-Glimmer rewrites for pools of 12 to 22 versions and 37.9\% at $k = 23$, and it became unstable across the random splits (Supplementary Section~S8.2).

The two screening scenarios thus accumulate different errors.
The union catches nearly every AI-rewritten abstract and falsely flags one in eight human-written abstracts.
The pooled detector keeps false flags near 1\% and misses one in three rewrites of Muse-Glimmer.
Calibrating the union as one screen brings its false flags down to the level of the pooled detector and keeps a median true positive rate of 99.3\% across the 23 versions, yet it still misses one in six of the Muse-Glimmer rewrites (Supplementary Section~S8.1).

\subsection{A case study of a deployed commercial detector}

The screening scenarios above rest on detectors that we trained ourselves on our corpus.
We now ask how a detector that institutions can actually deploy divides these two errors.
We examine Pangram~\cite{emi2024pangram,glickenhaus2026pangram}, a commercial detector of text written with an LLM.
NeurIPS, one of the leading conferences in artificial intelligence, partnered with Pangram to screen the submissions of its position paper track in 2026~\cite{lu2026neurips}.
PNAS Nexus, a sister journal of PNAS, lists Pangram among the tools used in its editorial assessment of submissions~\cite{pnasnexus2026screening}.
We sent 400 AI-rewritten abstracts per LLM version and the 5,000 calibration abstracts to Pangram through its API, with the same default production model, version 3.3.2, serving all requests (see Methods).

For a given text, Pangram returns one of three labels: `fully AI-generated', `AI-assisted, but not fully AI-generated', and `fully human-written'.
Of the 5,000 human-written calibration abstracts, it labels 4,999 as fully human-written and one as fully AI-generated.
The deployed detector thus showed a false positive rate of 0.02\% (1 of the 5,000 calibration abstracts; exact 95\% confidence interval 0.0005\% to 0.11\%), far below the 1\% at which our detectors operate.

\begin{figure}[t]
\centering
\includegraphics[width=1\linewidth]{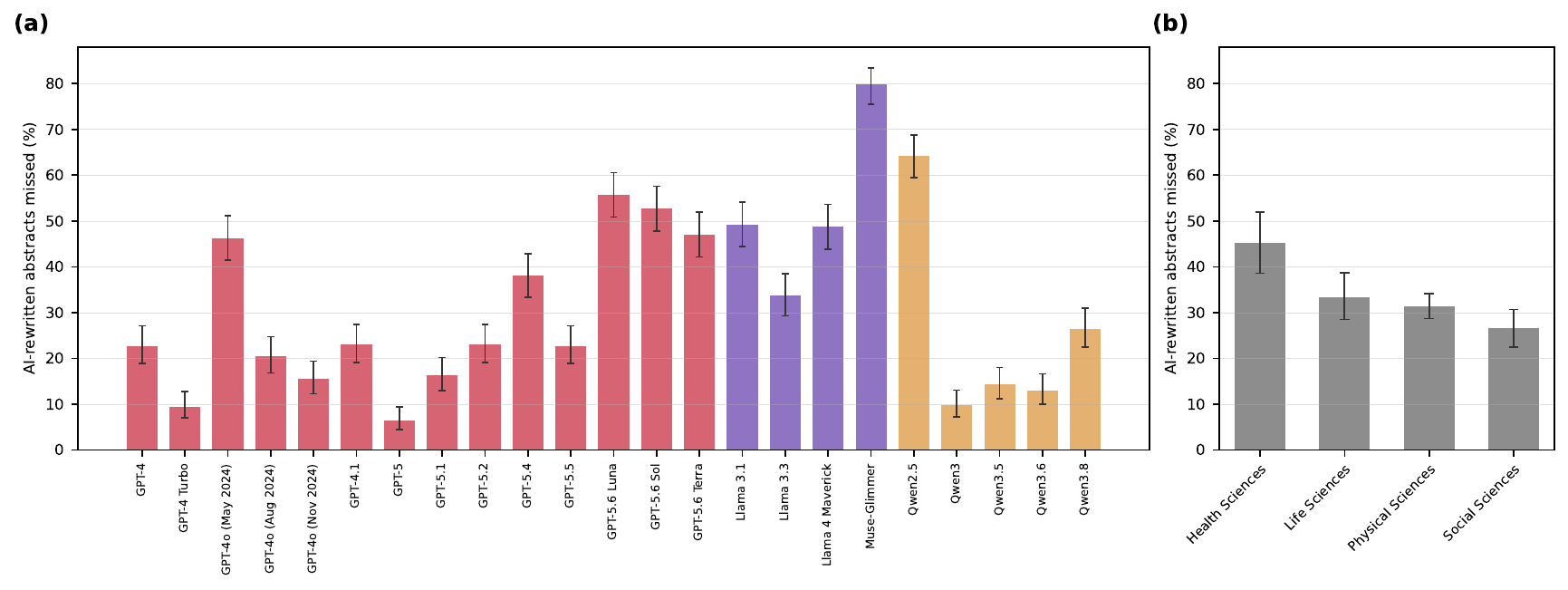}
\caption{
AI-rewritten abstracts missed by the deployed commercial detector.
(a) The share of AI-rewritten abstracts labeled fully human-written, for each of the 23 LLM versions, out of the same 400 papers per version.
(b) The same share by the disciplinary domain of the original abstract, across the 23 rewrites of each of the 400 papers.
The four domains are those assigned by OpenAlex.
Error bars show 95\% confidence intervals: Wilson intervals in (a), and paper-level bootstrap intervals in (b), where the 23 rewrites of one paper enter together.
}
\label{fig:4}
\end{figure}

In contrast, the share of AI-rewritten abstracts that Pangram misses depends strongly on the LLM version (see Fig.~\ref{fig:4}(a)).
We count an AI-rewritten abstract as missed only when Pangram labels it fully human-written, the choice most favorable to the detector.
The missed share ranges from 6.5\% (26 of 400) for GPT-5 to 79.8\% (319 of 400) for Muse-Glimmer.
The missed share is not significantly related to the release date, where the Spearman rank correlation between the two across the 23 versions is $\rho = 0.25$ ($p = 0.25$), and some late versions, such as GPT-5, are among the easiest to catch.
Moreover, the missed share differs across the disciplinary domains of the original abstracts (see Fig.~\ref{fig:4}(b)).
Across all AI-rewritten abstracts, Pangram misses 45.2\% in the Health Sciences, 33.4\% in the Life Sciences, 31.4\% in the Physical Sciences, and 26.5\% in the Social Sciences.
The misses of the deployed detector thus fall unevenly across disciplinary domains, as the flags of human-written text have been reported to do~\cite{liang2023biased,park2026style}.

The deployed detector thus errs on the same side as the pooled scenario, with even fewer false flags and even more misses.
At its single fixed operating point, it flagged 1 of the 5,000 human-written abstracts, and its misses reach four in five rewrites for Muse-Glimmer.

\section{Discussion}

Across 23 versions from three vendors, we quantified how detection of AI-rewritten scientific abstracts behaves as newer LLM versions arrive.
A detector trained on the version it screens catches almost all of its rewrites, whereas the same detector may miss most rewrites of a later generation from the same vendor.
Rather than accumulating gradually with model age, these drops concentrated at a few transitions between versions, which coincided with the largest vocabulary shifts.
The practical challenge for journals and conferences is therefore that screening has to be maintained against a set of models that keeps changing.

Benchmarks for the detection of LLM use in text commonly compare human-written texts and those generated or rewritten by the large language models available when the detector is built~\cite{li2024mage,wang2024m4,dugan2024raid,wu2024detectrl,yu2025cheat,liu2024detectability,liu2025generalization}.
Some of these benchmarks also score detectors on generators held out from the training data~\cite{li2024mage,dugan2024raid}, yet the pool of models stays the one assembled when the benchmark was built, without the release order in which versions replace one another.
EvoBench is a recent exception, as it evaluates detectors across successive versions of the same model family and, for supervised detectors, trains once on the earliest version of each family~\cite{yu2025evobench}.
Another recent study rewrote arXiv abstracts with multiple LLMs and measured how detectors trained on one of them perform on the others, classifying individual sentences at a score threshold of 0.5~\cite{ren2026moving}.
In contrast, we retrain a detector at each release, give it the versions released up to that point, and read the outcome at a fixed rate of falsely flagged human abstracts.
Our results show that the loss of detection concentrates at a few transitions rather than growing with the age of the model, a shape that these benchmarks leave open.

How are detectors of LLM use to be maintained in scientific publishing?
Whether a detector still applies should be indicated by the versions that have appeared since its last validation, rather than by the time elapsed since then.
Strategies that reduce the frequency of retraining, such as training once on the earliest version of a model family~\cite{yu2025evobench}, may hold for as long as no such model transition intervenes.
Methods that adapt a detector when a new generator arrives, whether through training-free adaptation~\cite{guo2024detective}, class-incremental learning~\cite{li2025continual}, lightweight transfer~\cite{sun2026ridgeft}, or test-time adaptation~\cite{ren2026moving}, address the cost of updating a detector.
Our results add the question of which LLM arrivals call for such an update, which in turn requires a signal that can be read from the new LLM version itself.

The vocabulary difference between the rewrites of a new LLM version and those of earlier versions, which we measured by the JSD, can be a practical early signal.
Note that it requires only a sample of rewrite texts by LLMs.
In our simulations, the JSD largely tracked where detection transferred and where it failed within and across the three vendors.
Previous studies related the transfer of detection across LLMs to model-side properties such as parameter size or the type of update that produced a model~\cite{antoun2024text,yu2025evobench}; however, sizes and architectures are often undisclosed for major LLM versions.
Distances between the embeddings of texts generated by different LLMs have also been reported to correlate with how well a detector trained on one LLM performs on another~\cite{ren2026moving}.
The JSD is therefore one candidate input for judging which newly released versions may require re-verification of the detectors in service, and its usefulness for future releases warrants further investigation.

Earlier LLM versions also remain within the scope of screening, because submissions and the published literature contain text written with many LLM versions at once.
In our simulations, a detector trained on a later version missed the rewrites of an earlier version of the same vendor nearly as often as the reverse; for 41 of the 104 pairs of versions from the same vendor, the detector trained on the later version caught fewer than half of the rewrites of the earlier one, and for 47 pairs the detector trained on the earlier version caught fewer than half of the rewrites of the later one (Supplementary Section~S6).
The detectors trained on the three latest OpenAI versions, GPT-5.6 Luna, Sol, and Terra, caught hardly any of the rewrites of GPT-4 (Section~2.1).
The JSD tracked this backward transfer as it did the forward one (Supplementary Section~S6).
Keeping earlier versions in the training data restored this coverage, at a cost: the pooled detector caught 98.5\% of the rewrites of the earlier Meta versions on average (Supplementary Table~S7), and missed one in three Muse-Glimmer rewrites, which the detector trained on Muse-Glimmer alone caught at 97.4\% (Fig.~\ref{fig:1}), and training it on all of the available rewrites did not remove this cost (Supplementary Section~S8.2).
Each update of a detector may therefore need to be validated on earlier versions as well, and the choice of which versions a detector keeps covering also decides whose text it is likely to miss.

Beyond the maintenance of detectors, screening distributes two kinds of error, flagging human-written abstracts and missing rewrites.
Detectors can flag human-written text at rates that depend on how its author writes, with higher rates for writing by non-native English speakers~\cite{liang2023biased} and with rates that shift when a manuscript has been professionally edited~\cite{park2026style}.
Screening at a low rate of flagging human-written texts has been recommended in the first place~\cite{krishna2023paraphrasing,yu2026peerreview}.
Miss rates for LLM rewrites are not uniform either; in our evaluation of a commercial detector, the share of rewrites labeled human-written differed across disciplinary domains.
Previous studies have concentrated on the side of flagging human-written texts~\cite{liang2023biased,park2026style,yu2026peerreview}, whereas the incidence of missing rewrites by LLMs across authors or disciplinary domains has not yet been examined in comparable detail.
Deciding between the two errors is therefore not only a matter of tuning a scientific publishing system but also distributes the cost of screening across those whose manuscripts pass through it.

Our results are obtained from one journal, one language, and one type of text, namely abstracts published in PNAS between 2009 and 2018 and written in English.
Although PNAS is a multidisciplinary journal whose abstracts pass through its editorial process, scientific writing elsewhere may follow more varied patterns.
The generality of our qualitative findings calls for further investigation.
Additionally, we simulated the use of an LLM in scientific writing as a two-stage rewrite of an abstract, following Refs.~\cite{liang2024mapping,liang2025quantifying}.
On the other hand, its use in practice ranges from correcting grammar to rewriting a text in its entirety~\cite{liu2024detectability,yu2025cheat} and can involve many exchanges between authors and LLMs~\cite{dycke2026yourtext}.
A systematic study across levels and modes of LLM assistance to scientific writing would show how far the behavior reported here extends.

Our study has additional limitations.
First, our simulations rely on one family of trained detectors, RoBERTa-base fine-tuned on our corpus, and on one commercial detector at its fixed operating point.
A word frequency-based detector reproduced the collapse at generational boundaries and its association with the JSD (see Supplementary Section~S7), and four publicly released detectors without in-domain training on our corpus missed most rewrites of every version (see Supplementary Section~S5), yet these cover only a few detector designs, and detectors built on other principles may behave differently.
Second, the 23 versions come from three vendors, namely every version of OpenAI's GPT-4 through GPT-5 line of text models and Meta's and Alibaba's versions since Llama 3.1 and Qwen2.5; other vendors, earlier releases of these three vendors, and the lighter parallel models that these vendors release alongside them are not represented.
Finally, the JSD is associated with the transfer of detection after controlling for release timing, vendor, and rewrite length (see Supplementary Section~S6), and this association does not establish causation.

Despite these limitations, our work shows that the reliability of screening for LLM use in scientific writing can collapse with a single model release.
The question for the scientific community is therefore not only how accurate its detectors are, but against which LLM versions that accuracy was last verified.

\section{Methods}

\subsection{Human-written corpus}

We construct a collection of 34,925 abstracts, hereafter the ``human-written corpus''.
The abstracts come from papers published in PNAS between 2009 and 2018.
Because this ten-year window ends four years before the public release of ChatGPT~\cite{openai2022chatgpt}, we assume that the corpus predates the era of AI-assisted scientific writing. 
We collected the papers from OpenAlex, a large-scale open bibliographic database~\cite{priem2022}, using its March 30, 2026 snapshot. 

We first retrieved all 42,062 records that OpenAlex indexes under the PNAS
venue identifier in this period and that carry a non-empty abstract.
OpenAlex stores an abstract not as plain text but as the list of positions at which each word occurs, and we rebuilt the text by placing every word back at its recorded positions.
We kept the 37,307 records that satisfied the following criteria, all based on the metadata that OpenAlex records for each publication: (i) the record is typed as ``article''; (ii) its DOI begins with the canonical PNAS prefix (``10.1073/pnas.''); (iii) it is not flagged as retracted; and (iv) its language is labeled as English. 
We then removed leftover typesetting markup, such as HTML tags for italics or superscripts, with a standard HTML parser (Beautiful Soup version 4.12 in Python).
We also decoded character entities and collapsed repeated whitespace.
After cleaning, we excluded every record that matched any of three deterministic rules. 
First, we excluded the 1,957 records whose abstract field begins with ``Proceedings of the National Academy of Sciences (PNAS), a peer reviewed journal''; in these records, the field holds journal boilerplate instead of the paper's abstract. 
Second, we excluded the 422 records whose abstract field begins with ``Significance'' or ``Author Summary''; in these records, the field holds the significance statement while the technical abstract is missing. 
Third, we excluded the 3 records whose abstract field holds a fragment of fewer than 15 words: the single word ``a'', the phrase ``International audience'', and a truncated erratum notice. 
The three rules matched disjoint sets of records; their union comprises $1{,}957 + 422 + 3 = 2{,}382$ records, and excluding them left the 34,925 abstracts of the human-written corpus. 

Each of the 34,925 abstracts carries one of four disciplinary domains assigned by OpenAlex: Health Sciences, Life Sciences, Physical Sciences, and Social Sciences. 
We randomly sampled 1,000 abstracts per domain using a fixed random seed, yielding 4,000 abstracts to be rewritten by LLMs.
We reserved the remaining 30,925 abstracts for calibrating detection thresholds, and no detector was trained on the papers they come from.

\subsection{AI-rewritten corpus} 

We focus on 23 LLM versions from three major vendors: OpenAI (14 versions), Meta (4 versions), and Alibaba (5 versions), released between June 2023 and August 2026.
We conservatively excluded Anthropic's Claude models in view of the vendor's requirements and our intended use of outputs from its API.
Each version is a separate release that we address by a fixed identifier, the API model name or the weight revision hash in Table~S1; we call them versions to emphasize that each belongs to a vendor's model series and succeeds earlier releases in that series.
We included every version of OpenAI's GPT-4 through GPT-5 line of text models that was released in this window, and the versions of the Meta and Alibaba series from Llama 3.1 and Qwen2.5 onward, following the selection rules of Supplementary Section~S1.
Every version remained individually accessible when we last verified availability on September 13, 2026, either under a version-specific model name through the vendor's official interface or as model weights that the vendor has published openly.

Rewriting or generating scientific abstracts with LLMs is an established way to build evaluation corpora for detectors~\cite{gao2023comparing,yu2025cheat,liu2024detectability,liu2025generalization,wang2024m4,wu2024detectrl}.
For each version, we rewrite each of the 4,000 abstracts with the same two-stage procedure, yielding one ``AI-rewritten corpus'' per version. 
The two-stage prompts are adopted from Refs.~\cite{liang2024mapping,liang2025quantifying}, where the model first compresses the original abstract into a list of its key points and then expands this list back into a single abstract-style paragraph. 
We added one passage to the second-stage prompt, which tells the model that the text is the abstract of a PNAS article, cites the journal's instruction to authors on length and audience (no more than 250 words, explaining the major contributions to the general reader)~\cite{pnas2026authors}, and asks for a single paragraph with nothing in the output except the abstract.
The prompts were otherwise identical, word for word, across all 23 versions, and their full text, with our addition marked, appears in Supplementary Section~S2. 
For the nine versions that we served ourselves (all Meta and Alibaba versions), we fixed the sampling temperature at 1.0 and capped each generation at 2,048 tokens.
For the six API versions that expose these controls (the OpenAI GPT-4-series versions), we applied the same temperature and token cap.
For the remaining eight OpenAI GPT-5-series versions, which accepted temperature only at its default value of 1.0 when we queried them, we used the provider defaults and capped each generation at 4,096 tokens.
The higher cap covers the tokens that these versions spend on internal reasoning.

We then applied the following two rules to the raw outputs. 
First, we discarded any output that was truncated at the token cap, empty, or a boilerplate refusal to perform the rewrite. 
Of the $23 \times 4{,}000 = 92{,}000$ requested rewrites, we discarded 47 (0.05\%): 38 outputs truncated at the token cap and 9 empty outputs. No output was a refusal.
This left 91{,}953 retained rewrites (Supplementary Table~S2).
Second, we removed conversational openings such as ``Here is a rewritten version:'' and closing remarks such as a word count from 349 of the 91{,}953 retained rewrites (0.4\%), using deterministic text patterns; these patterns change only the opening or closing text and leave the rewritten abstract itself untouched (Supplementary Section~S3).

The human-written abstracts reached us through publishing and indexing, which render typography in plain characters, for example, straight quotation marks, two hyphens for a dash, and subscript digits in parentheses, as in CO(2).
The rewrites, by contrast, kept the typography that the LLMs produce, such as curved quotation marks, dashes, and markup for italics and subscripts.
We then applied one deterministic normalization to every text that we used to train, calibrate, and evaluate our own detectors and to compute the JSD and the fidelity measures, human-written and AI-rewritten alike (see Supplementary Section~S3).
The normalization rewrites quotation marks, dashes, superscripts, subscripts, and markup into the plain forms of the human-written abstracts, and it changed 42,485 of the 91,953 rewrites (46.2\%) and 5,966 of the 34,925 human-written abstracts (17.1\%).
The texts that we submitted to the commercial detector and the rewrites that the two raters read are those before this normalization, with the typography that the LLMs produced.

\subsection{Trained detectors}

Our detectors are binary classifiers that label a given abstract as human-written or AI-rewritten.
We build each detector by fine-tuning RoBERTa-base, a pretrained language model with 125 million parameters~\cite{liu2019roberta}, on pairs of human-written and AI-rewritten abstracts.
Fine-tuning a pretrained classifier such as RoBERTa is a standard supervised approach in the detection of AI-generated text~\cite{solaiman2019,li2024mage,wang2024m4,wu2024detectrl,wu2025survey}.
Each abstract enters the detector as at most its first 512 tokens.
Training data come from the 3,955 of the 4,000 sampled papers whose rewrites were retained for all 23 versions. 
This restriction ensures that every detector is trained and evaluated on the same underlying papers. 
We split these papers into 80\% for training and 20\% for evaluation at the paper level, which guarantees that the human abstract and every rewrite of one paper fall on the same side of the split (3,164 training and 791 evaluation papers). 
Each detector is trained on the human-written abstract of each training paper and on one AI rewrite per training paper, which keeps the two classes balanced.
We fine-tune for 4 epochs with a batch size of 8 and the AdamW optimizer at a learning rate of $2 \times 10^{-5}$ with linear decay.
The detectors differ only in the set of model versions whose rewrites supply their training texts; the architecture, the training procedure, and every hyperparameter are held fixed across all experiments, except for the pooled detectors trained on all of the available rewrites (Supplementary Section~S8.2).
We repeat every experiment with 10 random splits and report means and standard deviations across the repetitions. 

\subsection{Detection scenarios}

We consider four maintenance scenarios that mirror how a detector could be kept up to date in practice. 
Each scenario specifies which LLM versions' rewrites train the detector that screens a target LLM version $v$.
In all four scenarios, the training versions come from the same vendor as $v$. 
In the ``current version'' scenario, the detector is trained on rewrites from $v$ itself; this idealizes a detector built with full access to the version being screened. 
In the ``past and current versions'' scenario, the detector is trained on rewrites from $v$ and from all versions released before $v$; this represents a detector retrained immediately whenever a new version appears. 
In the ``past versions only'' scenario, the detector is trained on rewrites from the versions released before $v$ and not from $v$ itself; this reflects the common situation that a new version's text arrives for screening before any training data for that version exist.
In the ``frozen at a specific version'' scenario, the detector is trained on rewrites from one designated version only and never updated afterward; this represents a legacy detector left in service without maintenance. 
In our experiments, the designated version is the earliest version we study from each vendor.

Beyond these four maintenance scenarios, each of which concerns one target LLM version at a time, we examine two screening scenarios that combine LLM versions, one at the decision level and one at the training level. 
In the ``union'' scenario, we run one detector per LLM version, each trained under the current version scenario, and flag an abstract as AI-rewritten when at least one of the 23 detectors does so; this represents an institution that stacks every available specialized detector in order not to miss any version. 
In the ``pooled'' scenario, we build a single classifier on rewrites pooled from all 23 versions; this represents the generalist design common in academic and commercial tools. 
To vary how many versions this design covers, we also train ``pooled($k$)'' classifiers that pool the first $k$ of the 23 versions in ascending order of release date, for $k$ from 2 to 23; pooled(23) is the pooled scenario itself. 

Whenever a scenario pools rewrites from several versions, the detector is still trained on one AI rewrite per training paper, drawn uniformly at random from the pooled versions; the draw is made once before training and stays fixed across epochs. 
The training sets are therefore matched paper by paper across scenarios: the same 3,164 training papers, the same 3,164 human abstracts, and one AI rewrite per paper, where only the LLM version that supplied each rewrite differs between scenarios.
This choice holds the number of rewrites used for training equal across every detector.
The pooled scenarios therefore measure detection under a fixed training budget.
We report pooled detectors trained on all of the rewrites available for the versions they cover in Supplementary Section~S8.2.

The last scenario is the one an institution can license: it deploys a commercial detector. 
We evaluate Pangram (version 3.3.2, the vendor's default model as of September 2026)~\cite{emi2024pangram}, which has been deployed for screening submissions to the NeurIPS 2026 conference~\cite{lu2026neurips} and the PNAS Nexus journal~\cite{pnasnexus2026screening}.
We query the product through its API, at its production setting and without access to its training data. 
We note that Pangram retired version 3.3.2 on September 30, 2026, after we had completed our queries.

Table~1 summarizes all detection scenarios examined in this study and the real-world situations they simulate.

\begin{table}[t]
\centering
\caption{
Detection scenarios examined in this study. Four maintenance scenarios that train one detector per target LLM version, two screening scenarios that combine versions, and one scenario in which a commercial detector is deployed. 
}
\label{tab:scenarios}
\small
\begin{tabular}{p{0.2\linewidth}p{0.33\linewidth}p{0.38\linewidth}}
\hline
Scenario & Definition & Real-world situation being simulated \\
\hline
Current version &
Trained on rewrites from the target version itself. &
An idealized detector built with full access to the exact version being
screened. \\ \addlinespace
Past and current versions &
Trained on rewrites from the target version and all versions released
before it. &
A detector retrained immediately whenever a new version appears. \\ \addlinespace
Past versions only &
Trained on rewrites from the versions released before the target version,
but not the target itself. &
The common situation: a new version's text arrives for screening before
any training data for that version exist. \\ \addlinespace
Frozen at a specific version &
Trained on rewrites from one designated version only and never updated (in
our experiments, the earliest version we study from the vendor). &
A legacy detector built once and left in service without maintenance. \\ \addlinespace
\hline
Union &
One detector per version, each trained under the current version scenario;
an abstract is flagged when at least one detector fires. &
An institution that stacks every available specialized detector so as not
to miss any version. \\ \addlinespace
Pooled &
A single classifier trained on rewrites pooled from all 23 versions. &
A generalist detector trained on a broad corpus of AI text, a design
common in academic and commercial tools. \\ \addlinespace
\hline 
Commercial detector &
A deployed product queried at its production setting, without access to
its training data. &
The screening service that a journal or institution would license. \\ 
\hline
\end{tabular}
\end{table}

\subsection{Thresholds and evaluation metrics}

Each detector that we train ourselves assigns a score to a given abstract, and a higher score indicates greater confidence that the abstract is AI-rewritten. 
Turning scores into binary decisions requires a threshold. 
The threshold determines the false positive rate (i.e., the share of human-written abstracts that score at or above it and are hence flagged as AI-rewritten). 
We compute one threshold per detector, using only the 30,925 calibration abstracts, which come from papers that no detector sees during training.
The union and pooled scenarios, whose false positive rates are themselves a main result of this study, use all 30,925 abstracts; all other experiments use a fixed random subsample of 5,000. 
Our primary operating point sets the false positive rate to 1\% by placing the threshold at the 99th percentile of the calibration scores, following the recommendation to evaluate detectors of AI-generated text at a fixed low false positive rate~\cite{krishna2023paraphrasing,tufts2025practical}; results at 5\% and 0.5\% appear in Supplementary Section~S7.
At a given operating point, we report the true positive rate (i.e., the share of AI rewrites among the evaluation papers that score at or above the threshold). 
In figures and tables, we write `TPR at 1\% FPR' for the true positive rate at the primary operating point. 
Union screening aggregates decisions instead of scores, where each of the 23 detectors turns its own score into a decision at its own threshold and the union flags an abstract when at least one detector does. 
For union screening, we therefore report the share of evaluation abstracts that at least one detector flags, separately for human abstracts and for AI rewrites.

The commercial detector requires its own accounting, because we evaluate the vendor-provided document-level verdict at the vendor's default setting, without selecting a decision threshold of our own.
For each of the 23 LLM versions, we submitted to the product's API the rewrites of the same 400 papers, a fixed set of the common papers, of which 40 are in the Health Sciences, 85 in the Life Sciences, 185 in the Physical Sciences, and 90 in the Social Sciences.
We also submitted the same 5,000 calibration abstracts as for our own detectors.
We measure the false negative rate as the share of the submitted rewrites that the product labels fully human-written; the other two labels both count in the detector's favor.
We measure the false positive rate as the share of the 5,000 calibration abstracts that the product does not label fully human-written.

\subsection{Validation of the rewrites}

We check in two layers whether AI rewriting substantially altered the content of the original abstracts. 
The aim is to detect serious loss or alteration of content, not to grade writing quality. 

The first layer computes four fidelity measures for all 91,953 retained rewrites: the cosine similarity between sentence embeddings of the original abstract and the rewrite~\cite{reimers2019sbert}, the share of numerals (e.g., integers, decimals, and percentages) in the original that reappear in the rewrite, the share of capitalized terms (a proxy for names and entities) that reappear in the rewrite, and the ratio of the rewrite's word count to the original's. 
Across the 23 versions, the rewrites stay semantically close to their originals, while a version-dependent share of numerals and capitalized terms disappears; Supplementary Section~S4 gives the exact definitions and the full distributions.

The second layer is a human check. 
Two raters, the first author and a research assistant, independently examined 80 rewrites: 40 papers, each rewritten by two versions chosen to span different vendors and both serving modes (GPT-5, accessed through the vendor API, and Llama 4 Maverick, served from public weights). 
The rewrites were presented to the raters in randomized order and without model names, and each rater recorded the systematic rewriting patterns they noticed. 
Supplementary Section~S4 lists the patterns with definitions and examples and notes which of them the two raters found independently of each other.
This check characterizes recurring changes in a sample of rewrites, and it does not estimate how often a rewrite alters the substance of its original across the 23 versions.

\section*{Acknowledgments}

K.N.~acknowledges financial support in part from JST ASPIRE Grant Number JPMJAP2328 and in part from JSPS KAKENHI Grant Number JP24K21056.
This work was supported by ABCI 3.0, which is provided by AIST and AIST Solutions.
While preparing this manuscript, the first author used ChatGPT (OpenAI) and Claude (Anthropic) for language editing, code development, and analysis assistance.
The first author reviewed all text, code, and results.

\section*{Declaration of Competing Interest}
The authors declare no competing interests.

\newpage

\begin{center}
\vspace*{12pt}
{\Large Supplementary Information for:\\
\vspace{12pt}
Large Language Model Turnover Undermines Screening for Artificial Intelligence-Assisted Scientific Writing
}
\vspace{12pt} \\
\end{center}

\setcounter{figure}{0}
\setcounter{table}{0}
\setcounter{section}{0}

\renewcommand{\thesection}{S\arabic{section}}
\renewcommand{\thefigure}{S\arabic{figure}}
\renewcommand{\thetable}{S\arabic{table}}
\renewcommand{\theequation}{S\arabic{equation}}

\begin{center}
\author{Kazuki Nakajima and Takayuki Mizuno}
\vspace{24pt} \\
\end{center}

\section{The 23 LLM versions}
Table~\ref{tab:s1} lists the 23 LLM versions we focus on in this study. 
For OpenAI, which we accessed through its API, we included every version-specific model identifier of the GPT-4 through GPT-5 line of text models that remained accessible as of September 13, 2026.
Parallel tiers (mini, nano, and pro variants), reasoning-specialized o-series models, task-specific models, and rolling aliases that the vendor updates in place were out of scope. 
For Meta and Alibaba, which publish weights openly, we included one model per generation of the mainline instruct series from Llama 3.1 and Qwen2.5 onward, in the size class that the vendor released consistently across these generations.
This size class is the 70-billion-parameter class for Llama 3.1 and 3.3, the Maverick model for Llama 4, the 72-billion-parameter class for Qwen2.5, and the roughly 30-billion-parameter dense class for Qwen3 through 3.8. 
We further required that the weights could be served with the vLLM inference engine in our serving environment.
We order the versions by the dates in Table~\ref{tab:s1}, which come from vendor metadata rather than public announcements, and the two need not coincide.
Within each vendor, these dates order the versions as their version numbers do, so only the interleaving of versions from different vendors depends on the exact dates.
For the OpenAI versions through GPT-5.5, the date is the one in the model identifier; the identifiers of the three GPT-5.6 versions carry a code name instead, and for these we use the date of the creation timestamp that the vendor's models endpoint reports.
For Meta and Alibaba, the date is the creation date of the Hugging Face repository that hosts the weights.
The three GPT-5.6 versions share the same date, and we place them in alphabetical order of their names.
In the maintenance scenarios, a version counts as earlier than another only when its date is earlier, and thus none of the three enters the past versions of another; the screening scenarios add them one at a time in this order.

\begin{table}[p]
\centering
\caption{
The 23 model versions. ``Identifier'' is the date-pinned API model name, or the Hugging Face repository and the revision hash of the weights we served (verified against the repositories on September 14, 2026). ``Date used for ordering'' is the date by which we place the versions in release order (see Section~S1).
``Generation settings'': Standard API = temperature 1.0 with a 2,048-token cap via the vendor API; Reasoning API = provider defaults with a 4,096-token cap; Self-served = served with vLLM at temperature 1.0 and a 2,048-token cap.
}
\label{tab:s1}
\resizebox{\textwidth}{!}{
\begin{tabular}{llll}
\toprule
Version & Identifier & Date used for ordering & Generation settings \\
\midrule
GPT-4 & \texttt{gpt-4-0613} & 2023-06-13 & Standard API \\
GPT-4 Turbo & \texttt{gpt-4-turbo-2024-04-09} & 2024-04-09 & Standard API \\
GPT-4o (May 2024) & \texttt{gpt-4o-2024-05-13} & 2024-05-13 & Standard API \\
GPT-4o (Aug 2024) & \texttt{gpt-4o-2024-08-06} & 2024-08-06 & Standard API \\
GPT-4o (Nov 2024) & \texttt{gpt-4o-2024-11-20} & 2024-11-20 & Standard API \\
GPT-4.1 & \texttt{gpt-4.1-2025-04-14} & 2025-04-14 & Standard API \\
GPT-5 & \texttt{gpt-5-2025-08-07} & 2025-08-07 & Reasoning API \\
GPT-5.1 & \texttt{gpt-5.1-2025-11-13} & 2025-11-13 & Reasoning API \\
GPT-5.2 & \texttt{gpt-5.2-2025-12-11} & 2025-12-11 & Reasoning API \\
GPT-5.4 & \texttt{gpt-5.4-2026-03-05} & 2026-03-05 & Reasoning API \\
GPT-5.5 & \texttt{gpt-5.5-2026-04-23} & 2026-04-23 & Reasoning API \\
GPT-5.6 Luna & \texttt{gpt-5.6-luna} & 2026-06-23 & Reasoning API \\
GPT-5.6 Sol & \texttt{gpt-5.6-sol} & 2026-06-23 & Reasoning API \\
GPT-5.6 Terra & \texttt{gpt-5.6-terra} & 2026-06-23 & Reasoning API \\
Llama 3.1 & \texttt{meta-llama/Llama-3.1-70B-Instruct@1605565b} & 2024-07-16 & Self-served \\
Llama 3.3 & \texttt{meta-llama/Llama-3.3-70B-Instruct@6f6073b4} & 2024-11-26 & Self-served \\
Llama 4 Maverick & \texttt{meta-llama/Llama-4-Maverick-17B-128E-Instruct-FP8@94125d2b} & 2025-04-01 & Self-served \\
Muse-Glimmer & \texttt{meta-models/Muse-Glimmer-30B@a4e59da5} & 2026-08-09 & Self-served \\
Qwen2.5 & \texttt{Qwen/Qwen2.5-72B-Instruct@495f3936} & 2024-09-16 & Self-served \\
Qwen3 & \texttt{Qwen/Qwen3-32B@9216db57} & 2025-04-27 & Self-served \\
Qwen3.5 & \texttt{Qwen/Qwen3.5-27B@fc05daec} & 2026-02-24 & Self-served \\
Qwen3.6 & \texttt{Qwen/Qwen3.6-27B@6a9e13bd} & 2026-04-21 & Self-served \\
Qwen3.8 & \texttt{Qwen/Qwen3.8-27B@1d4bf0f2} & 2026-08-05 & Self-served \\
\bottomrule
\end{tabular}}
\end{table}

\clearpage
\section{The two-stage prompts}
Each first-stage request consisted of the first-stage prompt below, followed by the line ``Text:'' and the original abstract. 
Each second-stage request consisted of the second-stage prompt below, followed by the line ``Concise version:'' and the model's first-stage output. 
The text shown in italics is our addition; everything else is taken unchanged from Refs.~\cite{S-liang2024mapping,S-liang2025quantifying}.

First-stage prompt:
\begin{quote}
The aim here is to reverse-engineer the author's writing process by taking a
piece of text from a paper and compressing it into a more concise form. This
process simulates how an author might distill their thoughts and key points
into a structured, yet not overly condensed form.\\
Now as a first step, first summarize the goal of the text, e.g., is it
introduction, or method, results? and then given a complete piece of text
from a paper, reverse-engineer it into a list of bullet points.
\end{quote}

Second-stage prompt:
\begin{quote}
Following the initial step of reverse-engineering the author's writing
process by compressing a text segment from a paper, you now enter the second
phase. Here, your objective is to expand upon the concise version previously
crafted. This stage simulates how an author elaborates on the distilled
thoughts and key points, enriching them into a detailed, structured
narrative.\\
Given the concise output from the previous step, your task is to develop it
into a fully fleshed-out text.\\
\emph{The text is the abstract of a research article in PNAS. Follow the
journal's instruction to authors: ``Please provide an abstract of no more
than 250 words in a single paragraph. Abstracts should explain to the
general reader the major contributions of the article.''\\
Output only the abstract itself, as one paragraph with no headings, bullet
points, or line breaks.}
\end{quote}

\newpage
\section{Discarding and formatting rules for the LLM outputs}
Table~\ref{tab:s2} accounts for the $23 \times 4{,}000 = 92{,}000$ requested rewrites. 
We identified refusals by flagging, for all 23 versions, outputs of fewer than 40 words that contained a refusal phrase such as ``I cannot''.  
No output was flagged as a refusal, and the 47 discarded outputs consist of 38 outputs truncated at the token cap and 9 empty outputs.

Some of the 91,953 retained rewrites opened with conversational text before the abstract itself, such as ``Here is a rewritten version:'', or ended with text after it, such as a word count. 
These passages are artifacts of the chat format rather than part of the rewritten abstract. 
We removed these passages with a fixed set of deterministic text patterns, applied identically to all 23 versions. 
A rewrite was edited only when its first characters matched one of the following opening phrases, compared case-insensitively: ``Here is a (rewritten) version/rewrite/abstract/expansion/summary'' and close variants, ``Sure,'', ``Certainly,'', ``Of course,'', ``Below is'', ``I'd be happy'', ``As requested'', or ``As an AI''. 
For a matching rewrite, we removed the text up to the first colon that is followed by a line break; when no such colon appeared within the first 200 characters, we removed the text up to the first line break, provided that the line break appeared within the first 200 characters; when neither appeared, the rewrite was left unchanged. 
In addition, we removed an opening that begins with ``Here is'', ``Here's'', or ``Here are'' and runs to a colon within its first line, a line of three or more hyphens at the start or the end of a rewrite, and a word count in parentheses at the end of a rewrite, such as ``(248 words)''.
The last column of Table~\ref{tab:s2} lists the number of edited rewrites per version. 
The edits concentrate in open-weight versions; for Llama~3.1, 222 of the 3,998 retained rewrites (5.6\%) opened or ended with such text. 
We kept the raw outputs unchanged alongside the edited corpus.

All texts that our own detectors and our measures of the rewrites use, the human-written abstracts and the AI rewrites alike, then went through one deterministic normalization.
The human-written abstracts reached us through publishing and indexing, which render typography in plain characters, whereas the rewrites kept the typography that the LLMs produce.
The normalization rewrites the following characters and markup into the plain forms found in the human-written abstracts, and it applies each rule in the same way to every text.
\begin{itemize}
\item HTML tags are removed with the same HTML parser as for the human-written abstracts, and escaped characters such as \texttt{\textbackslash x\{2019\}} are decoded.
\item The delimiters of inline LaTeX mathematics, such as \texttt{\$}$\ldots$\texttt{\$} and \texttt{\textbackslash(}$\ldots$\texttt{\textbackslash)}, are removed, and so is the backslash of LaTeX commands such as \texttt{\textbackslash alpha}; a dollar sign followed by a digit is kept as a currency sign.
\item Superscripts and subscripts of digits and charge signs are written as plain characters, so that CO(2) and CO\textsubscript{2} both become CO2 and 10\texttt{\^{}}16 becomes 1016; superscripts made of letters, such as Ly6C\texttt{\^{}}hi, are put in parentheses, as in Ly6C(hi).
\item Asterisks that mark a word or a phrase of up to six words in italics or bold, as in \texttt{*Aedes aegypti*}, are removed.
\item Characters with a compatibility form in Unicode, such as subscript and superscript digits, are replaced by that form (NFKC normalization).
\item Curved quotation marks and primes become straight quotation marks.
\item An em dash becomes two hyphens, and en dashes, minus signs, and nonbreaking hyphens become a hyphen.
\item The tilde operator (U+223C) becomes a tilde.
\item The placeholder ``[Formula: see text]'', which marks an omitted formula in some human-written abstracts, is removed.
\item Line breaks and runs of whitespace become a single space.
\end{itemize}
The normalization changed 42,485 of the 91,953 retained rewrites (46.2\%) and 5,966 of the 34,925 human-written abstracts (17.1\%), and it left no text empty.
Because the rules act on how characters are written, the frequency with which a version uses dashes or quotation marks, for example, is preserved.
The texts that we submitted to the commercial detector and the 80 rewrites of the human validation (Sections~\ref{section:s4.2} to \ref{section:s4.4}) are those before the normalization, so the counts of dashes in Table~\ref{tab:s3} refer to the typography that the LLMs produced.

\begin{table}[p]
\centering
\caption{
Accounting of the $23 \times 4{,}000 = 92{,}000$ requested rewrites. ``Refusals'', ``Truncated'', and ``Empty'' are the outputs discarded by the acceptance rules (47 in total; Methods), and ``Retained'' is the number of rewrites that remain out of the 4,000 requested per version. ``Openings or closings removed'' is the number of retained rewrites whose conversational opening, such as ``Here is a rewritten version:'', or closing, such as a word count, was removed by the formatting rules. The table lists every version with at least one discarded output or removed opening or closing; for the other 17 versions, all 4,000 rewrites were retained unchanged.
}
\label{tab:s2}
\small
\begin{tabular}{lrrrrr}
\toprule
Version & Refusals & Truncated & Empty & Retained & Openings or closings removed \\
\midrule
GPT-5             & 0   & 13 & 0  & 3,987 & 0 \\
Llama 3.1         & 0   & 2  & 0  & 3,998 & 222 \\
Llama 4 Maverick  & 0   & 0  & 0  & 4,000 & 88 \\
Muse-Glimmer      & 0   & 22 & 9  & 3,969 & 0 \\
Qwen3             & 0   & 0  & 0  & 4,000 & 35 \\
Qwen3.8           & 0   & 1  & 0  & 3,999 & 4 \\
\midrule
Total (23 versions) & 0 & 38 & 9 & 91,953 & 349 \\
\bottomrule
\end{tabular}
\end{table}

\clearpage
\section{Validation of the rewrites}

In this section, we check in two layers whether AI rewriting substantially altered the content of the original abstracts. 
The first layer computes four fidelity measures for all 91,953 retained rewrites (Section \ref{section:s4.1}). 
The second layer is a human check of a structured sample (Sections \ref{section:s4.2}--\ref{section:s4.4}).

\subsection{Distributions of the four fidelity measures}
\label{section:s4.1}

\begin{figure}[p]
\centering
\includegraphics[width=0.95\textwidth]{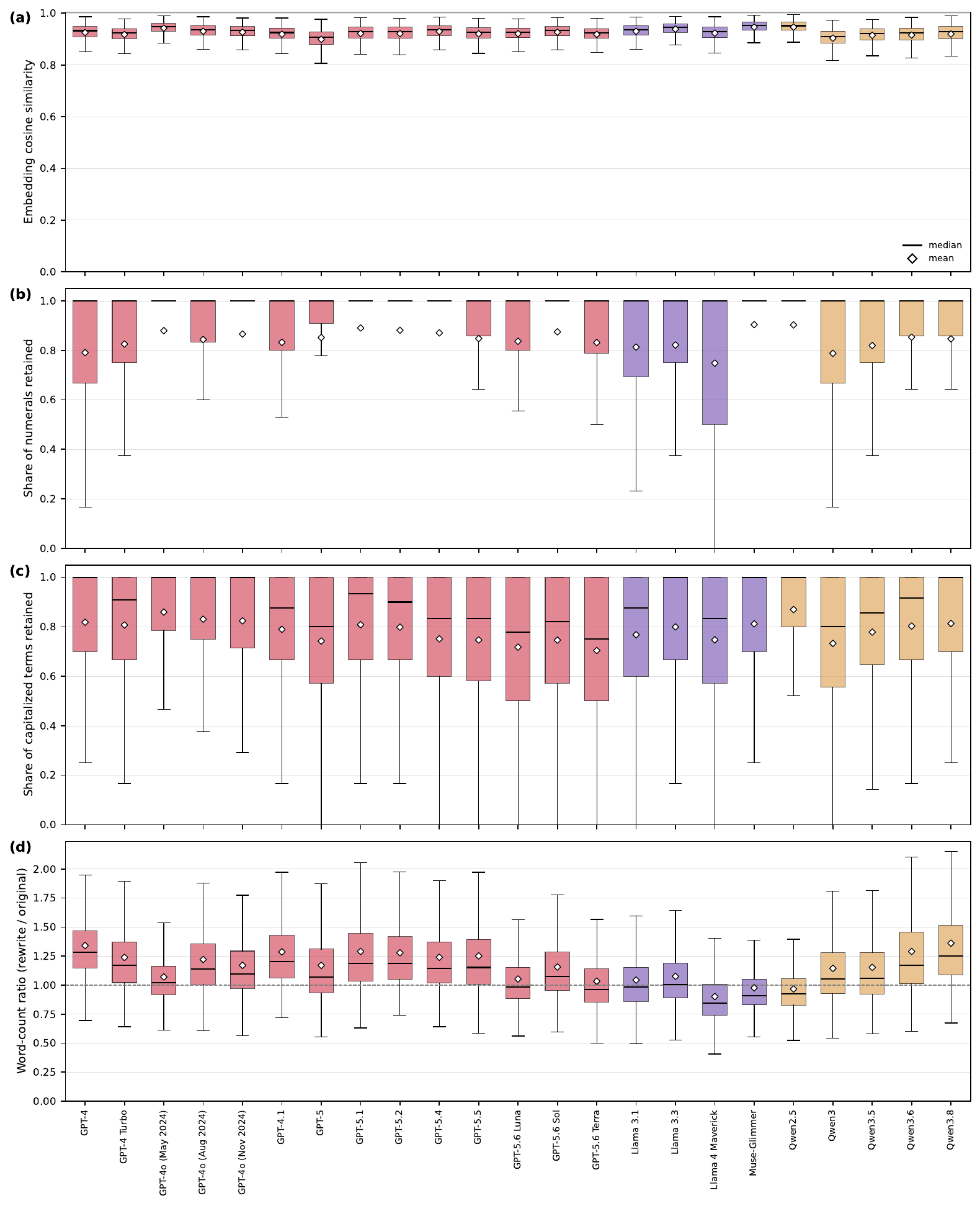}
\caption{
Distributions of the four fidelity measures for each of the 23 LLM versions.
(a) The embedding cosine similarity. (b) The share of numerals retained. (c) The share of capitalized terms retained. (d) The word-count ratio of the rewrite to the original.
Each box spans the interquartile range of the pairs of that LLM version, the black line inside the box is the median, and the white diamond is the mean.
The whiskers extend to the most extreme values within 1.5 times the interquartile range from the box, and values beyond the whiskers are not shown.
}
\label{fig:s1}
\end{figure}

We compare each rewrite with its original abstract along four measures.
Each measure targets one aspect of the correspondence between the two texts and yields one value per pair.
We compute the four measures for every pair of an original abstract and a retained rewrite, which gives 91,953 pairs in total across the 23 LLM versions.
Figure~\ref{fig:s1} shows the distribution of each measure for each LLM version.

The first measure is the semantic similarity between the rewrite and the original, as estimated by a sentence-embedding model.
We embed each original abstract and each rewrite into a 384-dimensional vector with the sentence-embedding model all-MiniLM-L6-v2~\cite{S-reimers2019sbert} (Hugging Face revision 1110a243, run with sentence-transformers 3.2.1) and compute the cosine similarity between the two normalized vectors.
We set the input window of the model to 512 tokens, which covers the full text of 99.8\% of the originals and 99.9\% of the rewrites.
For the remaining pairs, the model reads only the first 512 tokens of each text.
For each LLM version, we compute the mean and median of the cosine similarity over all of the pairs of that version.
This gives 23 means and medians, one per LLM version.
The smallest of the 23 means is 0.899 (GPT-5) and the largest is 0.946 (Muse-Glimmer and Qwen2.5).
The smallest of the 23 medians is 0.907 (GPT-5) and the largest is 0.953 (Muse-Glimmer).
For scale, 1,000 pairs of two different original abstracts drawn at random from the 4,000 originals yield a mean cosine similarity of 0.109 and a median of 0.084.

The second measure asks whether the numbers of the original reappear in the rewrite.
We extract from each text every numeral, defined as a match of the regular expression\\ \verb"(?<![A-Za-z0-9])(?:\d{1,3}(?:,\d{3})+|\d+)(?:\.\d+)?%?".
In words, a numeral is a run of digits, possibly with digit-grouping commas, a decimal part, and a trailing percent sign, and it must not start right after a letter or a digit.
Digit-grouping commas are then removed, and the numerals of a text form a set, in which duplicates count once.
For example, from the sentence ``CO2 levels rose by 3.5\% across 1,200 sites in 2020.'' this rule extracts the set \{3.5\%, 1200, 2020\}.
The measure is the share of the original's numerals that reappear in the rewrite.
When the original contains no numeral, the measure is set to 1.
Numerals appear in 2,707 of the 4,000 original abstracts (67.7\%), with a median of two distinct numerals per abstract among them.
The match is exact, and a numeral that reappears in a different form (for example, 0.37 rendered as 37\%) does not count as retained.
We compute the mean and the median of the measure over all of the pairs of each LLM version, as above.
The smallest of the 23 means is 0.748 (Llama 4 Maverick) and the largest is 0.904 (Muse-Glimmer).
All of the 23 medians equal 1, that is, for every LLM version at least half of the pairs retain every numeral of the original.

The third measure asks whether the names and technical terms of the original reappear in the rewrite.
We first split each text into sentences after every period, question mark, and exclamation mark followed by whitespace.
Within each sentence, we split the text into words at whitespace and remove from each word all characters other than letters, digits, and hyphens.
A word then counts as a capitalized term when it matches the regular expression \verb"^[A-Z][A-Za-z0-9-]{2,}$", that is, when it starts with a capital letter and has at least three characters.
The first word of a sentence counts only when it starts with at least two capital letters, which excludes ordinary sentence-opening words while keeping acronyms such as DNA.
For example, from the text ``DNA methylation increased. Notably, the Polycomb complex silenced HOX genes.'' this rule extracts the set \{DNA, Polycomb, HOX\}.
The capitalized terms of a text form a set, and the measure is the share of the original's terms that reappear in the rewrite.
When the original contains no capitalized term, we set the measure to 1.
Capitalized terms appear in 3,451 of the 4,000 original abstracts (86.3\%), with a median of five distinct terms per abstract among them.
We compute the mean and the median of the measure over all of the pairs of each LLM version.
The smallest of the 23 means is 0.704 (GPT-5.6 Terra) and the largest is 0.870 (Qwen2.5).
The smallest of the 23 medians is 0.750 (GPT-5.6 Terra) and the largest is 1 (eight versions).

The fourth measure is the length ratio, defined as the number of words of the rewrite divided by the number of words of the original.
A word is a maximal run of characters between whitespace.
This ratio does not measure content preservation on its own.
It flags rewrites that heavily compress or expand the original.
We compute the mean and the median of the ratio over all of the pairs of each LLM version.
The smallest of the 23 means is 0.901 (Llama 4 Maverick) and the largest is 1.361 (Qwen3.8).
The smallest of the 23 medians is 0.843 (Llama 4 Maverick) and the largest is 1.282 (GPT-4).
Overall, 63.7\% of the rewrites are longer than their originals.

\subsection{Independent checks by two raters} \label{section:s4.2}

In addition to the four measures above, two human raters read a sample of the rewrites against the original abstracts.
We drew the sample from the papers for which both GPT-5 and Llama 4 Maverick produced a retained rewrite.
Because these two LLM versions come from different vendors and are served in different modes, the sample covers two contrasting rewriting behaviors.
From these papers, we sampled 10 papers uniformly at random from each of the four disciplinary domains.
Each of the 40 sampled papers contributes its original abstract and its two rewrites, which gives 80 rewrites in total.

Each rater worked through the 40 papers in a browser tool prepared by the first author.
For each paper, the tool showed the original abstract and the two rewrites side by side, labeled A and B.
Which LLM version appeared as A was randomized paper by paper, with each version appearing as A in exactly 20 papers, and the raters were not told which version was behind either label.
The raters were asked to record, in free form, any pattern that recurred across the rewrites, such as habits of vocabulary, sentence structure, organization, or tone.
The two raters worked independently and did not discuss their notes during the check.
The raters were the first author and a research assistant; the second rater was not involved in the design or analysis of this study and took part only in this check.

After both raters finished, the first author merged the two sets of notes.
The recurring differences fell into two kinds.
The first kind concerns single words and punctuation marks, such as a preference for a specific word or an increased use of a specific punctuation mark.
Patterns of this kind can be counted mechanically, and Supplementary Section~\ref{section:s4.3} therefore reports exact counts over the 80 rewrites.
The second kind concerns the content of the rewrites, such as added interpretation or dropped details.
Supplementary Section~\ref{section:s4.4} organizes the patterns of this kind into a list with definitions and examples.

\subsection{Word and punctuation patterns} \label{section:s4.3}

We fixed 14 word and punctuation items.
Each item names a direction of change that a rater had noted (for example, the rewrites use the word `we' less often than the original, or the rewrites contain more hyphens).
For each item, we then counted the relevant unit in every original abstract and every rewrite of the sample.
For word items, the unit is the number of occurrences of the word family, counted without regard to letter case (for example, `show', `shown', and `showed' together).
For punctuation items, the unit is the number of occurrences of the character.
For the length item, the unit is the number of words.
A rewrite counts as showing an item when its count moves from the count of the original strictly in the noted direction.

The second rater also judged, for each of the 80 rewrites and each of the 14 items, whether the rewrite shows the item.
These judgments were made by hand, independently of the machine counting.
The machine counts reproduce the hand judgments in 1,087 of the 1,120 decisions (97.1\%), where the 1,120 decisions are the 14 items for each of the 80 rewrites.
Seven of the 14 items agree in all 80 rewrites.
The disagreements concentrate in the items that require counting many units by hand, with the lowest agreement at 68 of 80.

Table~\ref{tab:s3} lists the 14 items, the counted units, and the number of rewrites that show each item, separately for the two LLM versions.
Some items are specific to one LLM version.
For example, em dashes increase almost only in the GPT-5 rewrites (13 of 40, against 0 of 40 for Llama 4 Maverick), and the word `we' decreases mainly in the Llama 4 Maverick rewrites (28 of 40, against 7 of 40 for GPT-5).
Other items appear at similar rates in both LLM versions, such as a decrease of the word family of `show' (14 and 15 of 40).

\subsection{Content patterns} \label{section:s4.4}

We organized the content-level notes into 11 patterns.
Table~\ref{tab:s4} gives, for each pattern, a short name, a definition, and an example from the sample.
Each pattern describes a recurring way in which a rewrite differs from its original beyond single words and punctuation.
Three of the 11 patterns appeared independently in the notes of both raters, namely the use of an abbreviation without its definition, changes of notation, and the loss of specific details.
The remaining eight patterns appeared in the notes of one rater.

\begin{table}[p]
\centering
\caption{
The 14 word and punctuation items.
Each entry is the number of rewrites, out of 40 per LLM version, for which the stated pattern holds in comparison with the original abstract.
Words are counted without regard to letter case, the words of a family are counted together, and a pattern holds only when the count changes strictly in the stated direction.
}
\label{tab:s3}
\begin{tabular}{p{0.62\textwidth}rr}
\toprule
Pattern & GPT-5 & Llama 4 Maverick \\
\midrule
The rewrite contains more words than the original. & 21 & 7 \\ \addlinespace
The rewrite uses the word `we' less often than the original. & 7 & 28 \\ \addlinespace
The rewrite uses the word `our' more often than the original. & 8 & 21 \\ \addlinespace
The rewrite uses the word family `show', `shown', and `showed' less often than the original. & 14 & 15 \\ \addlinespace
The rewrite uses the word `findings' more often than the original. & 13 & 14 \\ \addlinespace
The rewrite uses the word family `analyse', `analyze', and `analytical' more often than the original. & 2 & 1 \\ \addlinespace
The rewrite uses the word `however' less often than the original. & 9 & 9 \\ \addlinespace
The rewrite uses the word family `significant', `significantly', and `significance' less often than the original. & 14 & 8 \\ \addlinespace
The rewrite uses the word family `known' and `unknown' less often than the original. & 7 & 8 \\ \addlinespace
The rewrite contains more em dashes (---) than the original. & 13 & 0 \\ \addlinespace
The rewrite contains more hyphens (-) than the original. & 30 & 5 \\ \addlinespace
The rewrite contains more en dashes (--) than the original. & 20 & 0 \\ \addlinespace
The rewrite contains more double quotation marks, straight or curly, than the original. & 5 & 1 \\ \addlinespace
The rewrite contains fewer opening parentheses than the original. & 16 & 25 \\
\bottomrule
\end{tabular}
\end{table}

\begin{table}[p]
\centering
\caption{
The 11 content patterns.
The patterns were derived from the free-form notes of the two raters, who read rewrites by GPT-5 and Llama 4 Maverick against the original abstracts (see Section~\ref{section:s4.2}).
All quoted fragments are verbatim from the rewrites in the sample.
Quotations from the original abstracts are limited to technical terms, notation, and single words; longer passages of the originals are described without quotation.
}
\label{tab:s4}
\small
\begin{tabular}{p{0.16\textwidth}p{0.33\textwidth}p{0.43\textwidth}}
\toprule
Pattern & Definition & Examples \\
\midrule
Undefined abbreviation & The rewrite uses an abbreviation without the full name that the original introduces it with. & The original introduces ``mitochondrial superoxide dismutase (MnSOD)'' with its full name, whereas the rewrite uses ``MnSOD'' throughout without the full name. \\
\midrule
Changed notation & Terms, units, symbols, letter case, or hyphenation are changed while the meaning is kept. & \textbullet~The original writes ``$\pi \to \pi^{*}$'' in symbols, whereas the rewrite spells it out as ``pi to pi*''. \newline \textbullet~The original writes ``200 mA/cm2'', whereas the rewrite writes ``200 mA cm$^{-2}$''. \\
\midrule
Dropped details & Specific information of the original, such as numbers, method names, species, or examples, is missing. & \textbullet~The everyday illustration that the original gives in parentheses does not reappear in the rewrite. \newline \textbullet~The method name ``Amino acid-coded mass tagging (AACT)'' of the original does not reappear in the rewrite. \\
\midrule
Added or altered content & Procedures, parameters, mechanisms, or numbers that the original does not state are added as if factual, or the object of a statement is changed. & \textbullet~The original attributes a finding to Eurasians as a whole, whereas the rewrite narrows it to ``Europeans and East Asians''. \newline \textbullet~The rewrite describes a titration procedure that the original does not mention. \\
\midrule
Added background & Textbook-style background or general statements are added. & The rewrite expands ``the hippocampus'' of the original into ``the hippocampus, a brain region critical for learning and memory''. \\
\midrule
Added significance & Significance, outlook, or implications absent from the original are added, typically at the end. & The rewrite adds the closing clause ``thereby shedding new light on the complex history of human migration and settlement''. \\
\midrule
Removed hedges & Cautious wording of the original is replaced by assertive wording. & The qualifier ``weakly'' that the original attaches to a reported increase does not reappear in the rewrite. \\
\midrule
Formulaic phrases & Recurring formulaic phrases appear across the rewrites. & \textbullet~The rewrite opens a sentence with ``To elucidate the underlying mechanisms, we investigated ...''. \newline \textbullet~The rewrite calls cancer ``a complex and multifaceted disease''. \\
\midrule
Changed person or voice & The person or the voice of the text changes systematically, including the removal of author statements of the ``Here we ...'' type. & The original reports the measurement in the first person, whereas the rewrite states ``magnetotransport measurements were conducted''. \\
\midrule
Reorganized structure & The order of sentences or the logical organization of the abstract is rebuilt. & The rewrite moves the main conclusion of the original to the opening sentences. \\
\midrule
Meta-reference & The rewrite refers to the abstract itself or adopts the tone of a third-party summary. & \textbullet~The rewrite describes itself with ``This perspective synthesizes how a new toolkit is overcoming that barrier''. \newline \textbullet~The rewrite reports a finding with ``as revealed by a study examining its expression and function''. \\
\bottomrule
\end{tabular}
\end{table}

\clearpage
\section{Detectors without training on our corpus}

In this section, we evaluate detectors that are not trained on our corpus.
We took four publicly released non-commercial detectors without any training on our corpus: the RoBERTa-based detector released by OpenAI in 2019 and trained on GPT-2 outputs~\cite{S-solaiman2019}, RADAR~\cite{S-hu2023radar}, Fast-DetectGPT~\cite{S-bao2024fastdetectgpt}, and Binoculars~\cite{S-hans2024binoculars}.
The first two are classifiers trained on the outputs of other LLMs (GPT-2 and Vicuna, respectively), and the last two are zero-shot detectors that score a text with fixed language models and require no training.

\begin{sloppypar}
We ran the four detectors on the same texts as the detectors trained on our corpus.
We used the released checkpoints: \texttt{roberta-base-openai-detector} for the OpenAI detector, \texttt{TrustSafeAI/RADAR-Vicuna-7B} for RADAR, the analytic sampling-free version of Fast-DetectGPT with \texttt{EleutherAI/gpt-neo-2.7B} serving as both the scoring and the reference model, and the official model pair \texttt{tiiuae/falcon-7b} and \texttt{tiiuae/falcon-7b-instruct} for Binoculars.
Each detector scored the 3,955 original abstracts shared by all 23 LLM versions, the 3,955 rewrites of each LLM version, and the 5,000 calibration abstracts.
Each detector reads at most the first 512 tokens of a text.
For each detector, we set the threshold at the 99th percentile of its scores on the 5,000 calibration abstracts, so that every detector operates at a false positive rate of 1\% on the calibration set, as in the main text.
A rewrite counts as detected when its score is at or above the threshold.
The released score of Binoculars is lower for machine text, and we therefore flipped its sign so that a higher score means machine text for all four detectors.
\end{sloppypar}

Table~\ref{tab:s5} lists the TPR at 1\% FPR of each of the four detectors for each of the 23 LLM versions.
All four detectors miss most rewrites.
Across the 23 LLM versions, the median TPR is 0.1\% for the OpenAI detector, 0.5\% for RADAR, 4.3\% for Fast-DetectGPT, and 2.5\% for Binoculars.
Even the best of the four detectors reaches a TPR of at most 30.3\% for any LLM version (Binoculars on Llama 4 Maverick) and below 10\% for 18 of the 23.
These misses do not come from the strict operating point alone.
Across the 23 LLM versions, the median AUROC is 0.55 for the OpenAI detector, 0.60 for RADAR, 0.67 for Binoculars, and 0.65 for Fast-DetectGPT.
The OpenAI detector stays below 0.65 for every version, and RADAR falls at or below 0.5, the value of a coin flip, for 9 of the 23 versions.
Where a detector does separate the two classes, the operating point still costs it.
The highest AUROC among all pairs of a detector and a version, 0.898 for RADAR on Llama 4 Maverick, comes with a TPR at 1\% FPR of 17.0\%.
These results are largely consistent with previous studies in that publicly released detectors fail to separate human-written text from AI-rewritten text in scientific writing~\cite{S-liang2025quantifying} and from AI-generated text at fixed low false positive rates~\cite{S-tufts2025practical}.

\begin{table}[p]
\centering
\caption{
True positive rate at a false positive rate of 1\% (in \%) of the four detectors without in-domain training, for each of the 23 LLM versions.
Each entry is the percentage of the 3,955 rewrites of that LLM version whose score is at or above the detector's threshold, rounded to one decimal; $<$0.1 marks a nonzero percentage below 0.05, and 0.0 marks that no rewrite was detected.
Each detector's threshold is set at the 99th percentile of its scores on the 5,000 calibration abstracts.
}
\label{tab:s5}
\begin{tabular}{lrrrr}
\toprule
LLM version & OpenAI detector & RADAR & Fast-DetectGPT & Binoculars \\
\midrule
GPT-4 & $<$0.1 & 3.1 & 0.1 & 0.1 \\
GPT-4 Turbo & 0.1 & 4.0 & 1.3 & 2.6 \\
GPT-4o (May 2024) & 0.4 & 6.5 & 4.4 & 7.2 \\
GPT-4o (Aug 2024) & 0.4 & 6.8 & 2.9 & 5.0 \\
GPT-4o (Nov 2024) & 0.2 & 2.9 & 8.5 & 9.4 \\
GPT-4.1 & 0.1 & 0.6 & 7.7 & 4.5 \\
GPT-5 & 0.0 & 0.0 & 0.1 & 0.0 \\
GPT-5.1 & 0.0 & $<$0.1 & 5.3 & 0.8 \\
GPT-5.2 & $<$0.1 & $<$0.1 & 0.8 & 0.0 \\
GPT-5.4 & 0.1 & 0.0 & 3.0 & 0.7 \\
GPT-5.5 & $<$0.1 & 0.0 & 3.7 & 0.5 \\
GPT-5.6 Luna & 0.1 & 0.2 & 4.5 & 2.0 \\
GPT-5.6 Sol & 0.1 & $<$0.1 & 1.5 & 0.3 \\
GPT-5.6 Terra & 0.1 & 0.1 & 2.2 & 1.0 \\
Llama 3.1 & 1.1 & 8.3 & 12.4 & 14.9 \\
Llama 3.3 & 1.6 & 12.7 & 16.8 & 20.1 \\
Llama 4 Maverick & 1.7 & 17.0 & 20.2 & 30.3 \\
Muse-Glimmer & 0.1 & 0.4 & 1.0 & 0.5 \\
Qwen2.5 & 1.6 & 9.1 & 19.8 & 26.5 \\
Qwen3 & $<$0.1 & 1.7 & 17.9 & 14.9 \\
Qwen3.5 & $<$0.1 & 0.2 & 1.2 & 0.6 \\
Qwen3.6 & 0.1 & 0.3 & 4.3 & 2.5 \\
Qwen3.8 & 0.2 & 0.5 & 7.6 & 3.4 \\
\bottomrule
\end{tabular}
\end{table}

\newpage
\section{Association between the JSD and detection}

We test whether the JSD between the rewrites of two LLM versions is associated with how well the detector trained on one of the two versions detects the rewrites of the other.
For a pair of LLM versions $(i, j)$, we take the TPR at 1\% FPR of the detector trained on the rewrites of version $i$ and evaluated on the rewrites of version $j$, averaged over the 10 random splits.
This gives a $23 \times 23$ matrix of TPRs.
The JSD is symmetric in the two versions of a pair, and we therefore compare it with the symmetric part of this matrix: for each of the 253 unordered pairs, we average the two TPRs of the pair, one for each train-to-test direction.
We measure the association by the Spearman rank correlation over the 253 pairs.
The 253 pairs are not independent observations, because they share only 23 underlying LLM versions.
We therefore assess significance with a Mantel permutation test~\cite{S-mantel1967detection}.
We permute the 23 version labels in the matrix of TPRs, recompute the correlation, and repeat this 10,000 times.
We take as the Monte Carlo $p$-value the share of permutations whose correlation is at least as extreme as the observed one, counting the observed one itself.

\begin{figure}[p]
\centering
\includegraphics[width=1\linewidth]{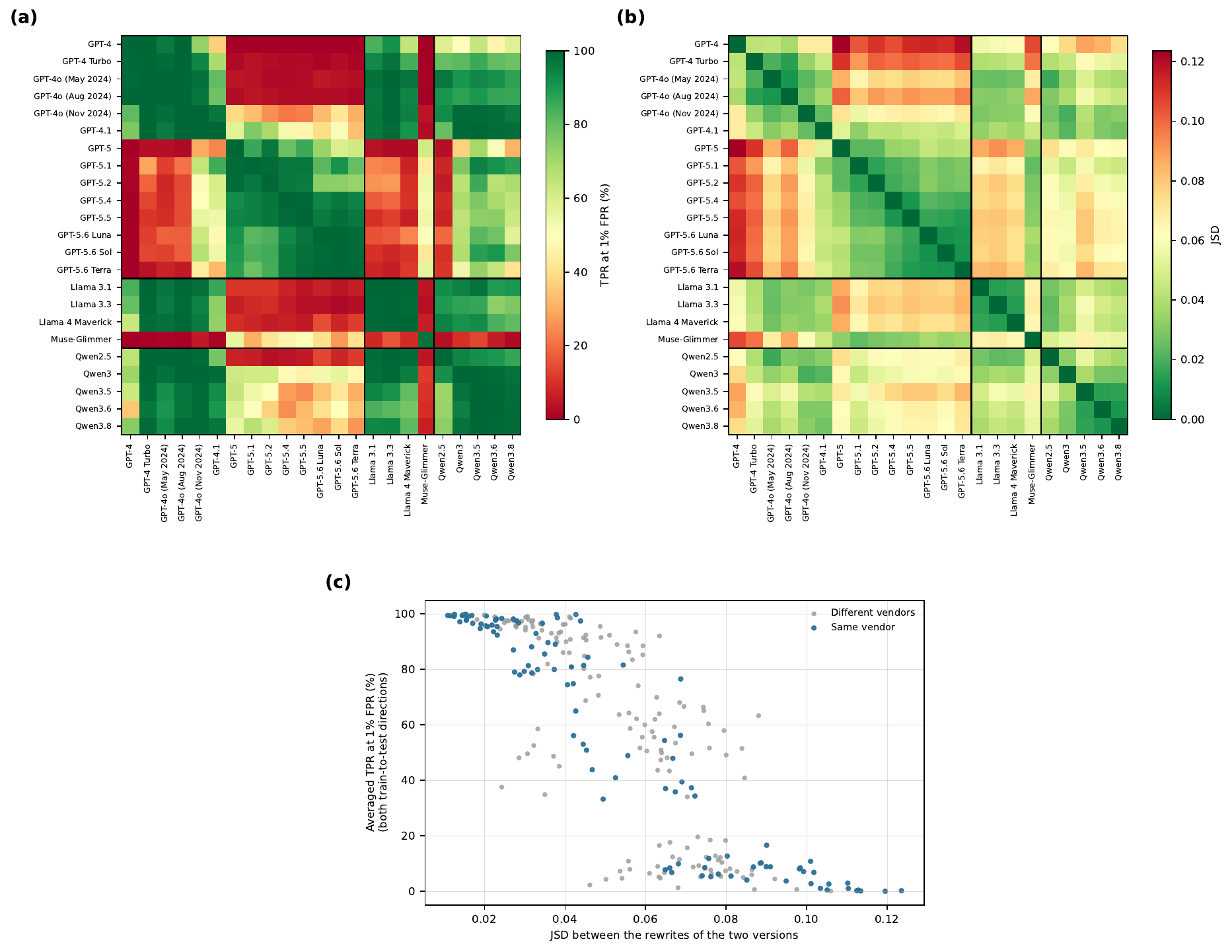}
\caption{
Association between the JSD and detection.
(a) Each cell shows the TPR at 1\% FPR of the detector trained on the rewrites of the row version and evaluated on the rewrites of the column version, averaged over the 10 random splits, on the color scale of the absolute value.
(b) Each cell shows the JSD between the rewrites of the row version and the rewrites of the column version, on the color scale of the absolute value; the diagonal equals zero.
(c) Each point is one of the 253 unordered pairs of LLM versions, with the JSD of the pair on the horizontal axis and the average of its two TPRs at 1\% FPR on the vertical axis.
}
\label{fig:s2}
\end{figure}

Figure~\ref{fig:s2} shows the TPRs and the JSDs of all pairs of versions and the scatter of the 253 unordered pairs behind the analyses of this section.
The correlation is $\rho = -0.85$, and no permutation reaches it, which gives $p < 0.001$.
Over the 250 pairs of versions released on different dates, the correlation is $-0.87$ in the direction from the earlier to the later version, and $-0.80$ in the reverse direction ($p < 0.001$ for both, by the same permutation test).
Within vendors, among the 104 pairs of versions released on different dates, the detector trained on the later version caught fewer than half of the rewrites of the earlier one for 41 pairs, and the detector trained on the earlier version caught fewer than half of the rewrites of the later one for 47 pairs (Fig.~\ref{fig:s2}(a)).
The association also holds separately within and across vendors, with $\rho = -0.92$ over the 107 pairs of versions from the same vendor and $\rho = -0.74$ over the 146 pairs of versions from different vendors ($p < 0.001$ for both, with the permutations restricted to shuffling the version labels only within each vendor).

We further check whether the association remains after controlling for three other properties of a pair: (i) the absolute difference of the dates in Table~\ref{tab:s1} in days, (ii) whether the two versions come from different vendors, and (iii) the absolute difference of the mean word counts.
The mean word count of an LLM version is the mean number of words over all of its retained rewrites.
We rank each of the five pair-level quantities over the 253 pairs, namely the averaged TPR, the JSD, and the three properties.
We then regress the ranked TPR and the ranked JSD on the three ranked properties and correlate the two residuals, which gives a partial Spearman correlation.
The partial correlation between the JSD and the TPR is $-0.80$.
Controlling for the date difference alone gives $-0.79$, and controlling for the vendor difference or the word-count difference alone leaves the correlation nearly unchanged ($-0.85$ and $-0.86$, respectively).

\clearpage
\section{Robustness checks}

In this section, we check whether the collapse of detection at generational boundaries (Fig.~1) and its association with the JSD (Fig.~2) persist under three changes of the analysis setup: another detector trained on our corpus, alternative prompts, and other operating points.

\subsection{Another detector trained on our corpus}

We build a word-frequency detector on the word-occurrence model of Ref.~\cite{S-liang2024monitoring}.
From the training data, it estimates, for each word $w$, the probability $p^{\mathrm{AI}}_{w}$ that $w$ occurs in a sentence of an AI rewrite and the probability $p^{\mathrm{H}}_{w}$ that $w$ occurs in a sentence of a human abstract.
We split each text into sentences with spaCy (en\_core\_web\_lg).
A word is a match of the regular expression \verb|\b\w+\b| in the lowercased sentence, and matches that consist only of digits are discarded.
The vocabulary $V$ contains the words that occur in at least five sentences of the training human abstracts and at least five sentences of the training rewrites.
A sentence $s$ receives the log-likelihood ratio under this model
\begin{align}
\mathrm{LLR}(s) = \sum_{w \in V} \left[ x_{w}(s) \log \frac{p^{\mathrm{AI}}_{w}}{p^{\mathrm{H}}_{w}} + \bigl(1 - x_{w}(s)\bigr) \log \frac{1 - p^{\mathrm{AI}}_{w}}{1 - p^{\mathrm{H}}_{w}} \right],
\label{eq:s1}
\end{align}
where $x_{w}(s) = 1$ if $w$ occurs in $s$ and $x_{w}(s) = 0$ otherwise.
The vocabulary restriction keeps every estimated probability away from zero, and the rare words whose estimated probability equals one, for which the log in Eq.~(S1) diverges, are discarded from $V$.
The score of a text is the logistic function applied to the mean of $\mathrm{LLR}(s)$ over the sentences of the text.
While this word-occurrence model was used to estimate the share of AI-modified text in a corpus of documents in Ref.~\cite{S-liang2024monitoring}, we instead use it to classify individual abstracts.

We repeat every experiment with this detector in place of the RoBERTa-based detector, with the same 3,955 common papers, the same 10 random splits, and the same evaluation rules.
The thresholds of all detectors in this section are set at the 99th percentile of their scores on the 5,000 calibration abstracts.

\begin{figure}[p]
\centering
\includegraphics[width=1\linewidth]{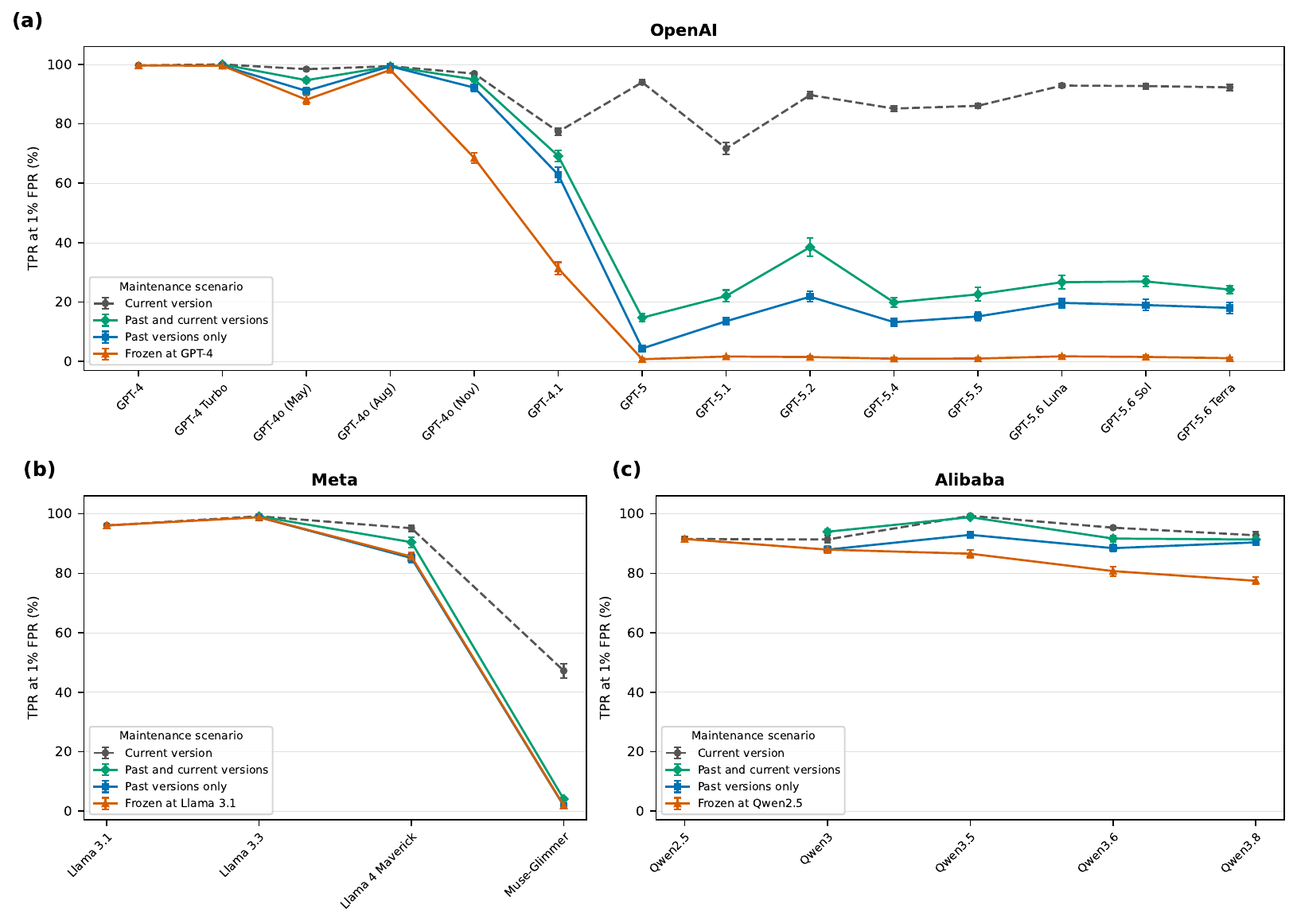}
\caption{
The four maintenance scenarios of Figure~1 of the main text, repeated with the word-frequency detector of Supplementary Section~S7.1.
Each panel shows one vendor: (a) OpenAI, (b) Meta, and (c) Alibaba.
The horizontal axis lists the vendor's LLM versions in release order, and the vertical axis shows the TPR at 1\% FPR.
Each line is one maintenance scenario, with the frozen detector trained on the earliest version we study from each vendor.
Error bars show the standard deviation over the 10 random splits.
}
\label{fig:s3}
\end{figure}

Figure~\ref{fig:s3} repeats the four maintenance scenarios of Figure~1.
The overall level is lower than in the main text: across the 23 LLM versions, the median TPR of the current-version detectors is 92.9\%, against 99.9\% for the RoBERTa-based detectors.
The collapse at generational boundaries is unchanged.
Trained on past versions only, the detector catches 4.4\% $\pm$ 0.6\% of the GPT-5 rewrites and 1.8\% $\pm$ 0.4\% of the Muse-Glimmer rewrites.

Over the 253 unordered pairs of LLM versions, the Spearman rank correlation between the JSD and the averaged TPR is $\rho = -0.83$, against $-0.85$ in the main text ($p < 0.001$ by the Mantel permutation test of Supplementary Section~S6).

\clearpage
\subsection{Alternative prompts}

The results presented in the main text come from one prompt, the two-stage prompt of Supplementary Section~S2 taken from Refs.~\cite{S-liang2024mapping,S-liang2025quantifying}.
In this section, we test whether the results of the main text depend on this choice.
We repeat the analyses behind Figures~1 and 2 with two alternative prompts.
Both come from the study of AI use in manuscript preparation of Ref.~\cite{S-chemaya2024perceptions}.
The first alternative prompt is the prompt `Rewrite 1b' of that study.
Each request consists of the following text, where the part in italics is our addition and is identical to the addition in Supplementary Section~S2:
\begin{quote}
Rewrite this paragraph in an academic language: ``[the original abstract]''\\
\emph{The text is the abstract of a research article in PNAS. Follow the
journal's instruction to authors: ``Please provide an abstract of no more
than 250 words in a single paragraph. Abstracts should explain to the
general reader the major contributions of the article.''\\
Output only the abstract itself, as one paragraph with no headings, bullet
points, or line breaks.}
\end{quote}
The second alternative prompt is the prompt `Rewrite 2' of that study without its closing sentence, which the authors state they appended for length control and which they also remove to form their `1b' variants:
\begin{quote}
Improve the clarity and coherence of my writing: ``[the original abstract]''\\
\emph{The text is the abstract of a research article in PNAS. Follow the
journal's instruction to authors: ``Please provide an abstract of no more
than 250 words in a single paragraph. Abstracts should explain to the
general reader the major contributions of the article.''\\
Output only the abstract itself, as one paragraph with no headings, bullet
points, or line breaks.}
\end{quote}
Both prompts consist of one stage.
We adopt only the prompt text from Ref.~\cite{S-chemaya2024perceptions}.
As everywhere in our corpus, the requests contain no system prompt, whereas that study used the system prompt ``You are a helpful assistant''.

For each of the 23 LLM versions and each alternative prompt, we generate rewrites of 500 original abstracts, with the same generation settings as in Table~S1.
The detector architecture, the training recipe, and the calibration abstracts are those of the main text, and each detector's threshold is recalibrated with the same procedure, at the 99th percentile of its scores on the calibration abstracts.
For each prompt, we keep the papers whose rewrites were retained for all 23 versions, which yields 498 papers for the first prompt and 494 papers for the second.
The 10 random splits divide these papers into 80\% training and 20\% test papers, and all reported values are means over the 10 splits.
Because this check works at about one eighth of the scale of the main text, we compare the alternative prompts with the two-stage prompt of the main text at the same scale; for this comparison, we restricted the main corpus to the same 500 original abstracts, retained the 495 papers whose rewrites were kept for all 23 versions, and repeated the analyses with 396 training papers, against 3,164 in the main text.

Under all three prompts at this scale, the detector trained on past versions only caught a much smaller share of the rewrites of GPT-5 and Muse-Glimmer than the detector trained on the version itself (Table~\ref{tab:s7}).
We recomputed the JSD from the rewrites of each prompt, and its association with the averaged TPR remained under both alternative prompts ($\rho = -0.66$ and $-0.73$, $p < 0.001$ for both, by the Mantel permutation test of Supplementary Section~S6), against $\rho = -0.81$ for the two-stage prompt at the same scale.

At this scale, detection was weaker overall than in the main text.
The median TPR of the current-version detectors across the 23 versions was 97.3\% for the two-stage prompt, 88.1\% for the first alternative prompt, and 75.9\% for the second, with the lowest values at 74.3\%, 54.5\%, and 36.6\%, respectively.

\begin{table}[p]
\centering
\caption{
Detection under the two-stage prompt of the main text and the two alternative prompts.
The first column repeats the main text, and the other three columns use the same 500 original abstracts.
All TPRs are at 1\% FPR and are means over the 10 random splits.
For the first version after each of the two generational boundaries, the second and third rows give the TPR of the detector trained on the version itself and that of the detector trained on past versions only.
The last row gives the Spearman rank correlation between the JSD and the averaged TPR over the 253 unordered pairs of versions, with $p < 0.001$ for all four by the Mantel permutation test of Supplementary Section~S6.
}
\label{tab:s7}
\small
\begin{tabular}{p{4.6cm}rrrr}
\toprule
 & \multicolumn{2}{c}{Two-stage prompt} & First & Second \\
 & Main text & 500 abstracts & alternative & alternative \\
\midrule
Papers retained for all 23 versions & 3,955 & 495 & 498 & 494 \\
TPR of current-version detectors, median (range) over 23 versions & 99.9 (97.4--100.0) & 97.3 (74.3--100.0) & 88.1 (54.5--99.9) & 75.9 (36.6--99.3) \\
GPT-5: current version $\to$ past versions only & 100.0 $\to$ 26.6 & 99.0 $\to$ 10.2 & 93.7 $\to$ 5.8 & 95.1 $\to$ 7.9 \\
Muse-Glimmer: current version $\to$ past versions only & 97.4 $\to$ 3.8 & 84.6 $\to$ 2.6 & 77.1 $\to$ 15.3 & 75.3 $\to$ 7.1 \\
Spearman $\rho$ between JSD and averaged TPR & $-0.85$ & $-0.81$ & $-0.66$ & $-0.73$ \\
\bottomrule
\end{tabular}
\end{table}

\clearpage
\subsection{Other operating points}

The results presented in the main text come from the operating point at a false positive rate of 1\%.
In this section, we repeat the analyses behind Figures~1 and 2 at two other operating points, 5\% and 0.5\% FPR.
We keep every trained detector and every score of the main text and only change the threshold to the 95th and the 99.5th percentile of the detector's scores on the calibration abstracts.

\begin{figure}[p]
\centering
\includegraphics[width=1\linewidth]{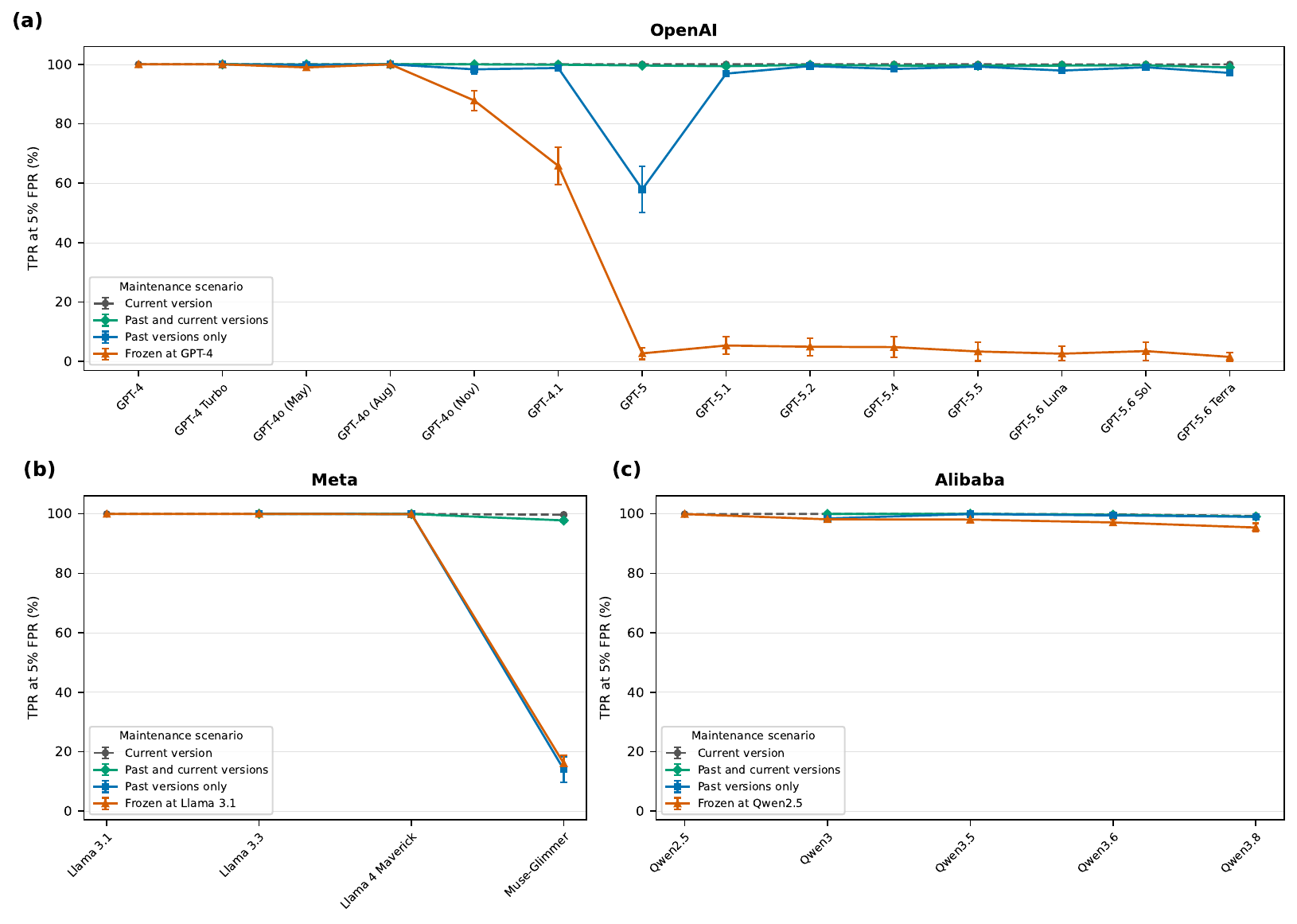}
\caption{
The four maintenance scenarios of Figure~1 of the main text, with every threshold set for a false positive rate of 5\% on the calibration abstracts.
Each panel shows one vendor: (a) OpenAI, (b) Meta, and (c) Alibaba.
The horizontal axis lists the vendor's LLM versions in release order, and the vertical axis shows the TPR at 5\% FPR.
Each line is one maintenance scenario, with the frozen detector trained on the earliest version we study from each vendor.
The detectors and their scores are those of the main text, and only the threshold is moved.
Error bars show the standard deviation over the 10 random splits.
}
\label{fig:s4}
\end{figure}

\begin{figure}[p]
\centering
\includegraphics[width=1\linewidth]{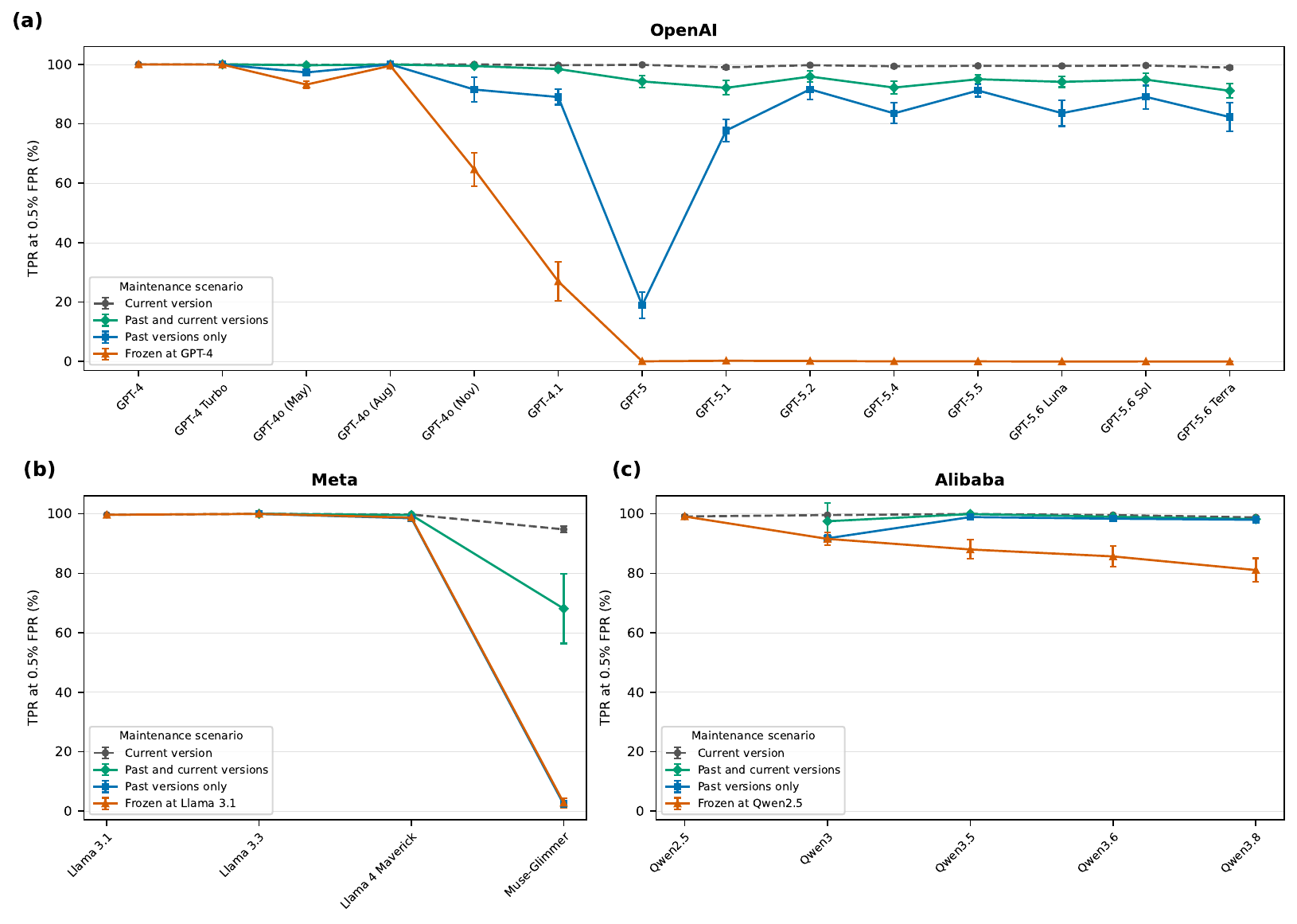}
\caption{
The four maintenance scenarios of Figure~1 of the main text, with every threshold set for a false positive rate of 0.5\% on the calibration abstracts.
Each panel shows one vendor: (a) OpenAI, (b) Meta, and (c) Alibaba.
The horizontal axis lists the vendor's LLM versions in release order, and the vertical axis shows the TPR at 0.5\% FPR.
Each line is one maintenance scenario, with the frozen detector trained on the earliest version we study from each vendor.
The detectors and their scores are those of the main text, and only the threshold is moved.
Error bars show the standard deviation over the 10 random splits.
}
\label{fig:s5}
\end{figure}

Figures~\ref{fig:s4} and \ref{fig:s5} repeat the four maintenance scenarios of Figure~1 at the 5\% and 0.5\% operating points, respectively.
The collapse at generational boundaries persists at both.
Trained on past versions only, the detectors catch 57.9\% $\pm$ 7.8\% of the GPT-5 rewrites and 14.0\% $\pm$ 4.3\% of the Muse-Glimmer rewrites at 5\% FPR, and 19.0\% $\pm$ 4.3\% and 2.2\% $\pm$ 1.2\% at 0.5\% FPR.

Over the 253 unordered pairs of LLM versions, the Spearman rank correlation is $\rho = -0.85$ at 5\% FPR and $\rho = -0.85$ at 0.5\% FPR, against $-0.85$ in the main text ($p < 0.001$ for both, by the Mantel permutation test of Supplementary Section~S6).

\newpage
\section{Additional results for the screening scenarios}

\subsection{Calibrating the union as one screen}

In the union scenario of Figure~3 of the main text, each of the 23 detectors carries its own threshold at the 99th percentile of its calibration scores, and the screen as a whole flags 11.9\% $\pm$ 1.2\% of the held-out human-written abstracts.
The pooled detector is a single detector and is calibrated as one, and it flags 0.8\% $\pm$ 0.3\%.
The two screens of Figure~3 are therefore compared at different rates of flagging human-written abstracts.
In this section we calibrate the union as one screen and repeat the comparison.
We convert each detector's score into its percentile within that detector's calibration scores, take the maximum of the 23 percentiles as the score of the screen, and place the threshold at the 99th percentile of that score on the calibration abstracts.

Calibrated in this way, the union flags 1.1\% $\pm$ 0.4\% of the held-out human-written abstracts, against 0.8\% $\pm$ 0.3\% for the pooled detector.
Table~\ref{tab:s6} gives the true positive rate of both screens for each of the 23 versions.
The two reach a similar level overall, with a median of 99.3\% for the union and 97.2\% for the pooled detector.
The largest differences are for Muse-Glimmer (83.6\% against 66.2\%), Qwen2.5 (97.7\% against 91.2\%), and GPT-4 (99.9\% against 94.4\%).
Muse-Glimmer is the version that both screens catch least, and the calibrated union misses one in six of its rewrites.

Calibrating the union as one screen thus brings its false flags down from 11.9\% to 1.1\%, and in exchange it misses more of the rewrites, most of all those of Muse-Glimmer (16.4\% against 1.4\% with a threshold per detector).

\begin{table}[p]
\centering
\caption{
True positive rate (in \%) of the three screens covering all 23 LLM versions, for each version.
The union flags an abstract when at least one of the 23 version-specific detectors flags it.
In the first column each detector uses its own threshold at the 99th percentile of its calibration scores, which is the setting of Figure~3 of the main text.
In the second column the union is calibrated as one screen, as described above.
The third column is the pooled detector trained on the rewrites of all 23 versions.
The last row gives the share of the held-out human-written abstracts that each screen flags.
Each entry is the mean $\pm$ the standard deviation over the 10 random splits.
}
\label{tab:s6}
\begin{tabular}{lrrr}
\toprule
LLM version & Union, per detector & Union, whole screen & Pooled \\
\midrule
GPT-4 & 100.0 $\pm$ 0.0 & 99.9 $\pm$ 0.1 & 94.4 $\pm$ 1.2 \\
GPT-4 Turbo & 100.0 $\pm$ 0.0 & 100.0 $\pm$ 0.0 & 100.0 $\pm$ 0.1 \\
GPT-4o (May 2024) & 100.0 $\pm$ 0.0 & 99.7 $\pm$ 0.2 & 98.4 $\pm$ 0.6 \\
Llama 3.1 & 100.0 $\pm$ 0.1 & 99.6 $\pm$ 0.2 & 98.4 $\pm$ 0.4 \\
GPT-4o (Aug 2024) & 100.0 $\pm$ 0.0 & 99.9 $\pm$ 0.1 & 99.7 $\pm$ 0.2 \\
Qwen2.5 & 99.9 $\pm$ 0.1 & 97.7 $\pm$ 0.5 & 91.2 $\pm$ 1.2 \\
GPT-4o (Nov 2024) & 100.0 $\pm$ 0.0 & 99.9 $\pm$ 0.1 & 99.8 $\pm$ 0.2 \\
Llama 3.3 & 100.0 $\pm$ 0.0 & 100.0 $\pm$ 0.0 & 99.8 $\pm$ 0.1 \\
Llama 4 Maverick & 99.9 $\pm$ 0.1 & 99.4 $\pm$ 0.2 & 97.4 $\pm$ 0.9 \\
GPT-4.1 & 100.0 $\pm$ 0.0 & 98.7 $\pm$ 0.3 & 97.2 $\pm$ 0.8 \\
Qwen3 & 100.0 $\pm$ 0.0 & 99.4 $\pm$ 0.3 & 99.4 $\pm$ 0.3 \\
GPT-5 & 100.0 $\pm$ 0.0 & 99.4 $\pm$ 0.3 & 95.8 $\pm$ 1.2 \\
GPT-5.1 & 99.9 $\pm$ 0.1 & 96.4 $\pm$ 1.0 & 93.8 $\pm$ 1.2 \\
GPT-5.2 & 100.0 $\pm$ 0.0 & 99.0 $\pm$ 0.3 & 95.5 $\pm$ 0.9 \\
Qwen3.5 & 100.0 $\pm$ 0.0 & 100.0 $\pm$ 0.1 & 99.9 $\pm$ 0.1 \\
GPT-5.4 & 99.9 $\pm$ 0.1 & 98.2 $\pm$ 0.6 & 94.8 $\pm$ 1.4 \\
Qwen3.6 & 99.8 $\pm$ 0.1 & 99.3 $\pm$ 0.3 & 98.9 $\pm$ 0.5 \\
GPT-5.5 & 100.0 $\pm$ 0.1 & 98.5 $\pm$ 0.4 & 95.6 $\pm$ 1.4 \\
GPT-5.6 Luna & 100.0 $\pm$ 0.1 & 98.3 $\pm$ 0.6 & 95.2 $\pm$ 1.4 \\
GPT-5.6 Sol & 100.0 $\pm$ 0.0 & 99.1 $\pm$ 0.3 & 95.3 $\pm$ 1.5 \\
GPT-5.6 Terra & 99.9 $\pm$ 0.1 & 97.8 $\pm$ 0.6 & 93.4 $\pm$ 1.8 \\
Qwen3.8 & 99.3 $\pm$ 0.1 & 98.6 $\pm$ 0.4 & 97.7 $\pm$ 0.5 \\
Muse-Glimmer & 98.6 $\pm$ 0.4 & 83.6 $\pm$ 2.8 & 66.2 $\pm$ 3.4 \\
\midrule
Human-written abstracts flagged & 11.9 $\pm$ 1.2 & 1.1 $\pm$ 0.4 & 0.8 $\pm$ 0.3 \\
\bottomrule
\end{tabular}
\end{table}

\clearpage

\subsection{Pooled detectors trained on all of the available rewrites}
The pooled detector of the main text, like every other detector in this study, is trained on one rewrite per training paper, drawn from the pooled versions (see Methods).
In this section, we train pooled detectors on all of the available rewrites of the pooled versions, to examine how much of the miss rate of the pooled detector depends on this fixed training budget.

For each $k$ from 1 to 23, we train a detector on the human-written abstracts of the 3,164 training papers and on the rewrites of the same papers by every one of the first $k$ versions in release order, that is, on $3{,}164 \times k$ rewrites, using the same 10 random splits as in the main text.
Because the rewrites then outnumber the human-written abstracts by a factor of $k$, each mini-batch of eight texts contains four human-written abstracts and four rewrites; every rewrite enters once per epoch, and the human-written abstracts are reused in a separately shuffled order.
The optimizer, the learning rate, and the number of epochs are those of the main text.
To avoid ties at the threshold when probabilities round to one, we use the difference between the two output logits as the score and place the threshold at its 99th percentile on the 30,925 calibration abstracts.
None of the 230 detectors gave degenerate scores: across them, the standard deviation of the scores on the calibration abstracts ranged from 0.68 to 10.4, and the share of the calibration abstracts flagged ranged from 1.002\% to 1.006\%.

\begin{figure}[p]
\centering
\includegraphics[width=1\linewidth]{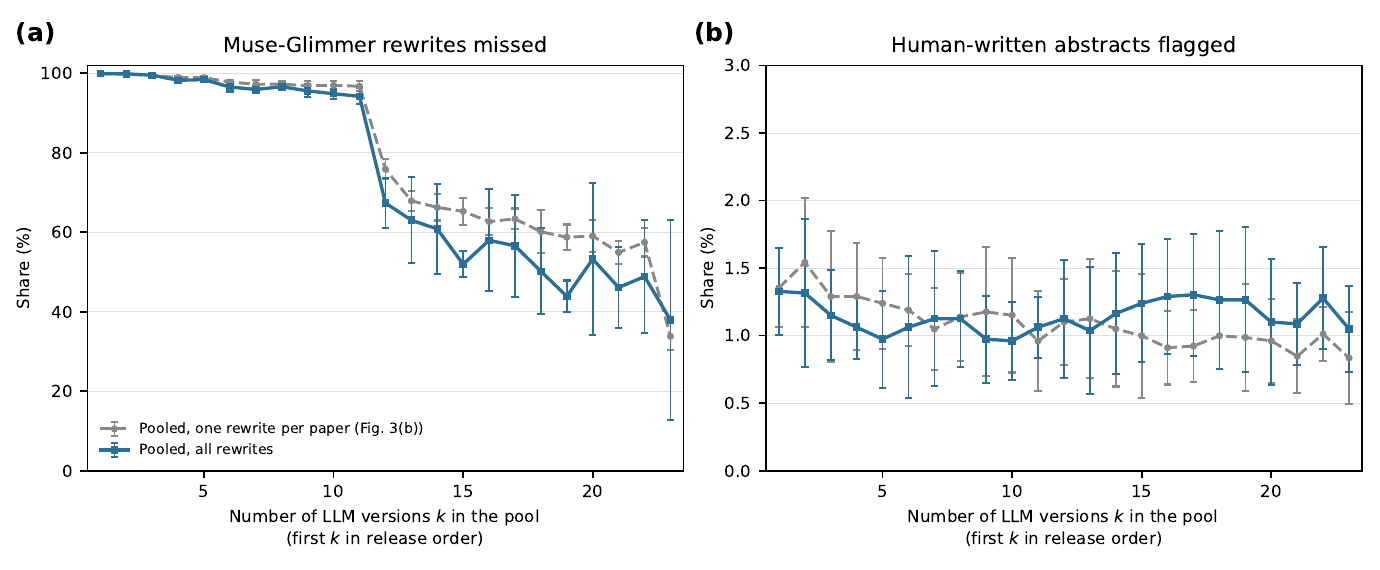}
\caption{
Pooled detectors trained on all of the available rewrites, compared with the pooled detector of the main text.
For each $k$, the detectors are built from the first $k$ LLM versions in release order.
(a) Share of the rewrites of the last released LLM version, Muse-Glimmer, that the detector misses.
(b) Share of the held-out human-written abstracts that the detector flags, on a different vertical scale.
Blue solid curves show detectors trained on all of the rewrites of the training papers by the first $k$ versions, and grey dashed curves show the pooled detector of the main text, trained on one rewrite per training paper (Figure~3(b) of the main text).
Points are means and error bars are standard deviations over the 10 random splits.
}
\label{fig:s6}
\end{figure}

Training on all of the available rewrites kept the false flags near 1\% for every $k$, with mean flagged shares of the held-out human-written abstracts between 0.96\% and 1.33\%.
It did not close the miss rate of the pooled detector on the rewrites of Muse-Glimmer (Fig.~\ref{fig:s6}(a)).
For pools of 12 to 22 versions, all released before Muse-Glimmer, it missed 5 to 15 percentage points fewer Muse-Glimmer rewrites than the pooled detector of the main text, and it still missed 44\% to 67\% of them; trained on all of the rewrites of the 22 earlier versions, it caught 51.1\% $\pm$ 14.3\% of them, against 42.5\% $\pm$ 3.6\%.
When the Muse-Glimmer rewrites entered the pool ($k = 23$), it missed 37.9\% $\pm$ 25.2\% of them (median 31.2\%), against 33.8\% $\pm$ 3.4\% for the pooled detector of the main text.
Across the 10 random splits at $k = 23$, it missed between 10.7\% and 74.7\% of the Muse-Glimmer rewrites; even the best split missed four times as many as the detector trained on Muse-Glimmer alone (2.6\%).
At the GPT-5 boundary, trained on all of the rewrites of the 11 versions released before GPT-5, it caught 35.6\% $\pm$ 4.5\% of the GPT-5 rewrites, against 26.7\% $\pm$ 5.5\%.

Training on all of the available rewrites was also unstable across the random splits.
In 22 of the 230 detectors, all with $k \geq 9$, the median TPR over the versions in the training data fell below 50\%, and in 18 of them below 12\%; these detectors missed most rewrites of the very versions they were trained on.
None of the 220 pooled detectors of the main text, trained on one rewrite per paper, failed in this way, and their lowest median TPR over the versions in the training data was 96.6\%.
The failing detectors did not give degenerate scores, and they flagged the held-out human-written abstracts at the same rate as the other detectors (1.16\% against 1.14\% on average); we therefore include them in the means and standard deviations above.

\clearpage

\renewcommand{\refname}{Supplementary References}

\end{document}